\documentclass{article}
\usepackage{authblk}
\usepackage{graphicx}
\usepackage{hyperref}
\usepackage{multirow}
\usepackage{rotating}
\usepackage{tabularx}
\usepackage{subcaption}
\usepackage{url}
\usepackage[colorinlistoftodos]{todonotes}

\begin{document}

\title{From Simulation to Real-World Robotic Mobile Fulfillment Systems}
	
\author[1]{Lin Xie\footnote{xie@leuphana.de}}
\author[2]{Hanyi Li}
\author[1]{Nils Thieme}
\affil[1]{Leuphana University of L\"uneburg, L\"uneburg, Germany}
\affil[2]{Beijing Hanning ZN Tech Co.,Ltd, Beijing, China}

\date{\today}
\maketitle

\begin{abstract}
In a new type of automated parts-to-picker warehouse system -- a Robotic Mobile Fulfillment System (RMFS) -- robots are sent to transport pods (movable shelves) to human operators at stations to pick/put items from/to pods. There are many operational decision problems in such a system, and some of them are interdependent and influence each other. In order to analyze the decision problems and the relationships between them, there are two open-source simulation frameworks in the literature, Alphabet Soup and RAWSim-O. However, the steps between simulation and real-world RMFS are not clear in the literature. Therefore, this paper aims to bridge this gap. The simulator is firstly transferred as core software. The core software is connected with an open-source ERP system, called Odoo, while it is also connected with real robots and stations through an XOR-bench. The XOR-bench enables the RMFS to be integrated with several mini-robots and mobile industrial robots in (removed) experiments for the purpose of research and education.
\end{abstract}

\section{Introduction}\label{sec:intro}
A Robotic Mobile Fulfillment System (RMFS) is a new type of automated parts-to-picker warehousing system, in which robots are used to transport movable shelves (also called pods), containing the inventory items, back and forth between the storage area and the replenishment or picking stations. Human operators work only at stations, either picking or replenishing items. In the former, items are picked from pods to fulfill customers’ orders, while items are refilled in pods in the latter. An RMFS aims to keep human workers at the stations busy while minimizing the resources (e.g. robots, stations, pods) to fulfill the incoming pick orders.

\begin{figure}[htb]
	\centering
	\begin{subfigure}[b]{0.33\textwidth}
		\includegraphics[width = \textwidth]{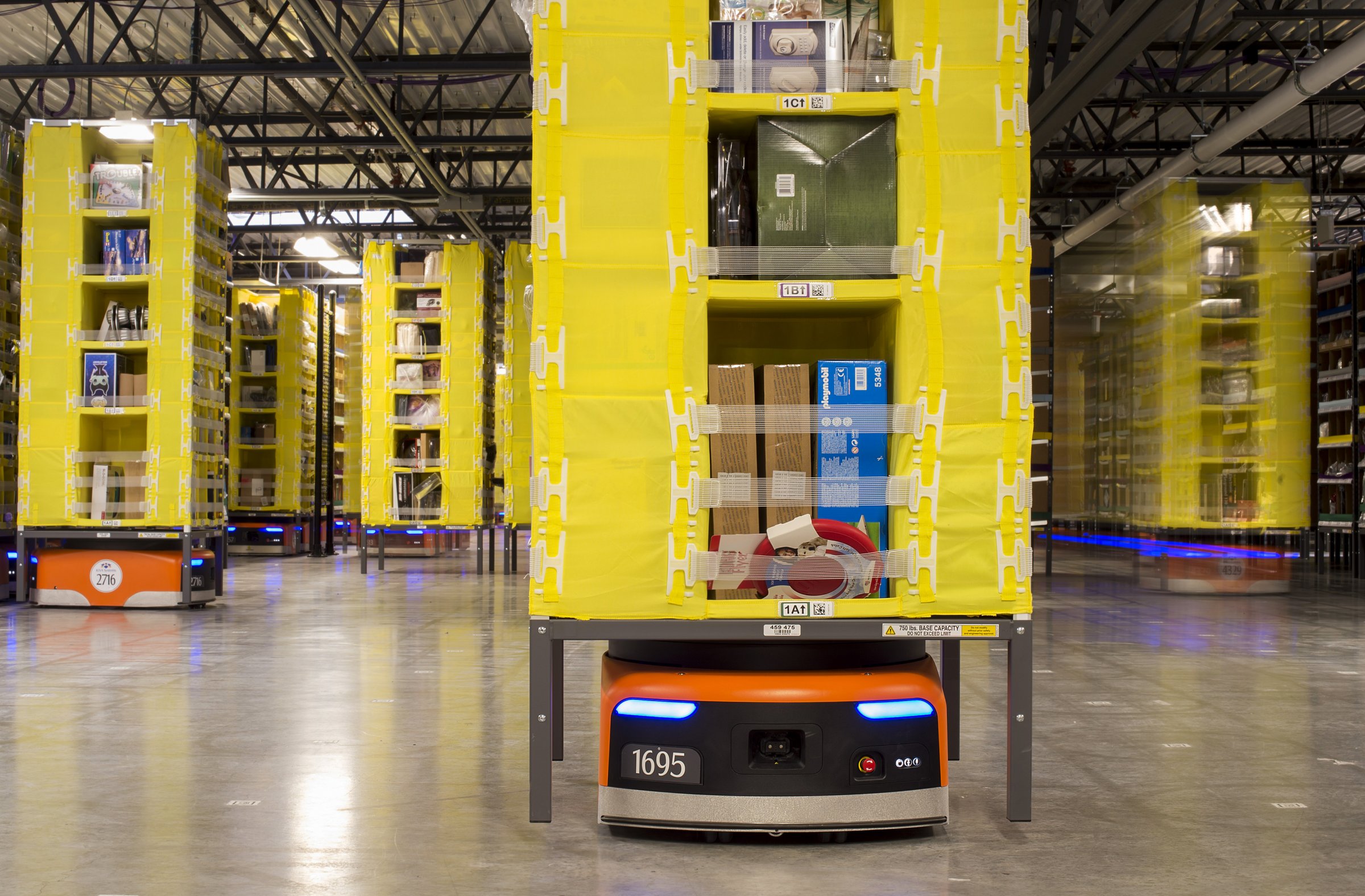}
		\caption{Amazon robots (Source: Amazon Robotics)}
		\label{fig:amazonrobots}
	\end{subfigure}
	~
	\begin{subfigure}[b]{0.3\textwidth}
		\includegraphics[width = \textwidth]{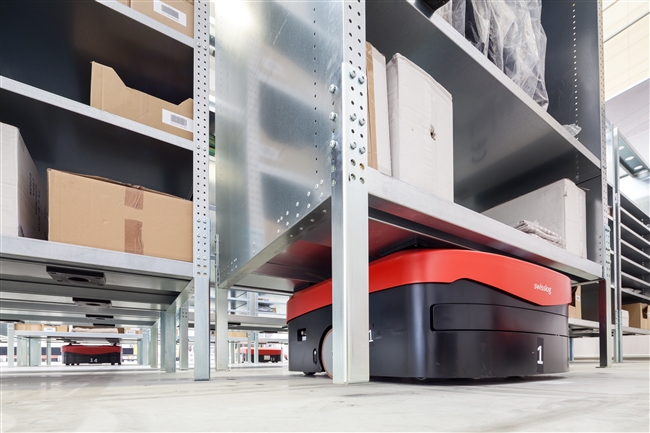}
		\caption{CarryPick\textsuperscript{TM} (Source: Swisslog, KUKA)}
		\label{fig:swisslog}
	\end{subfigure}
	~
	\begin{subfigure}[b]{0.3\textwidth}
		\includegraphics[width = \textwidth]{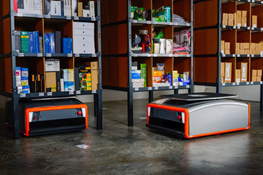}
		\caption{Butler\textsuperscript{TM} (Source: Grey Orange)}
		\label{fig:greyorange}
	\end{subfigure}
	~
	\begin{subfigure}[b]{0.3\textwidth}
		\includegraphics[width = \textwidth]{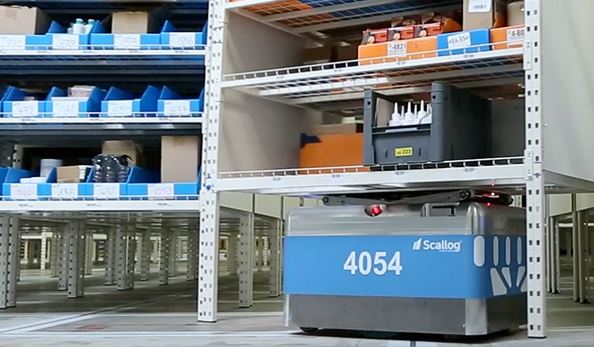}
		\caption{Scallog System\textsuperscript{TM} (Source: Scallog)}
		\label{fig:scallog}
	\end{subfigure}
	~
	\begin{subfigure}[b]{0.3\textwidth}
		\includegraphics[width = \textwidth]{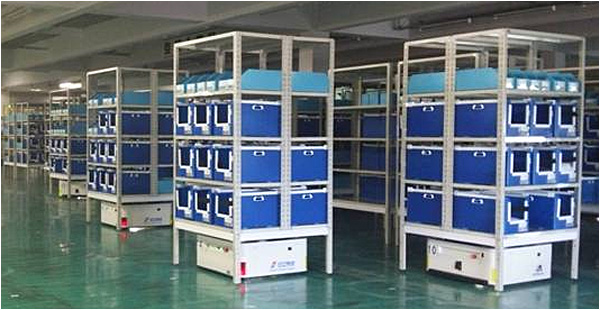}
		\caption{Racrew\textsuperscript{TM} (Source: Hitachi)}
		\label{fig:hitachi}
	\end{subfigure}
	\caption{Real-world RMFSs.}
	\label{fig:realworld}
\end{figure}

\begin{figure}[h]
	\centering
	\includegraphics[width=\textwidth]{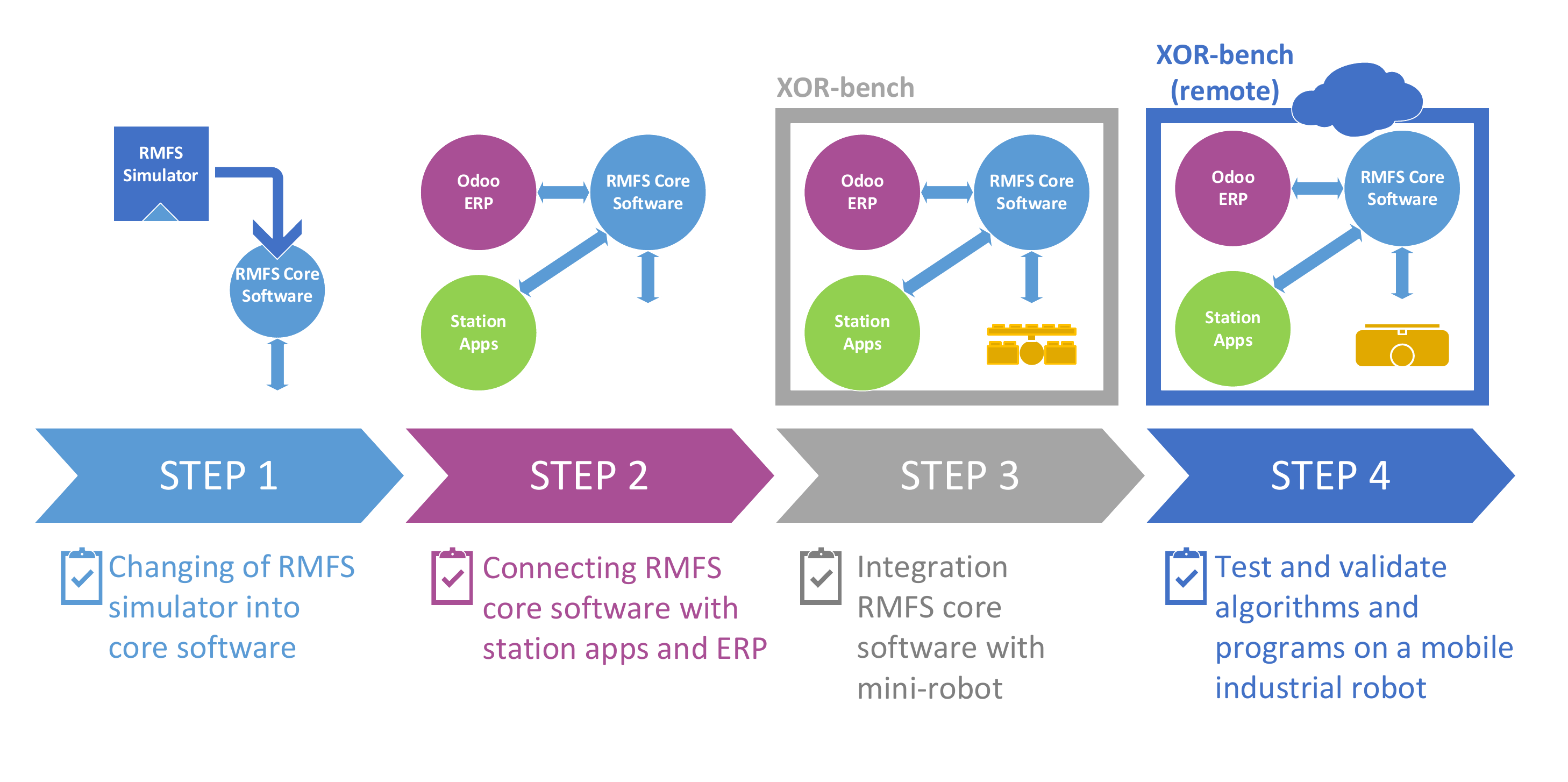}
	\caption{From simulation to real-world RMFS in four steps.}
	\label{fig:4_steps}
\end{figure}

RMFS has received more attention in the last decade, and some real-world RMFSs are currently under development (see Figure~\ref{fig:realworld}). There are numerous operational decision problems in this system, such as the decisions as to which robots will carry which pods to which station to fulfill an incoming order. Most of the research about such systems is focused on these problems and algorithms to improve system performance (see an overview in \cite{Merschformann-xie-li:2017} and \cite{de-koster-le-duc-roodbergen:2018}). Also, some of them are tested with a simulation framework (such as Alphabet Soup of \cite{Hazard.2006} or RAWSim-O of \cite{Merschformann-xie-li:2017}). This simulates the process of an RMFS, while it controls the resources. However, how can we make sure that the algorithms we implement and test in the simulator are still applicable for the real-world scenarios? This is the question we focus on in this paper. In order to answer this, four steps are required from simulation to real-world RMFS as illustrated in Figure \ref{fig:4_steps}. The first step is changing an RMFS simulator into RMFS core software. We use here our developed open-source simulator RAWSim-O. The second step is connecting the RMFS core software with an ERP system and station apps. The third step is the integration of the RMFS core software with mini-robots in an XOR-bench. The XOR-bench is a teleoperated platform for experiments. After the successful testing with mini-robots, we use the XOR-bench to test amd validate algorithms and programs remotely on a mobile industrial robot. The processes for these last two steps are highly complex and require interdisciplinary work between robotics engineering, software engineering and management science. 

Therefore, there are three main contributions in this work, which are summarized as follows:
\begin{itemize}
	\item We designed a process which aims to provide a seamless transition from the RMFS simulator to real-world RMFS.
	\item We connect the RMFS core software with an open-source ERP system (Odoo); moreover, the developed station apps are connected as well.
	\item We designed an XOR-bench, which aims to integrate the RMFS core software with the real robot and test and validate the programs on a mobile industrial robot.
\end{itemize} 

We describe the RMFS and RAWSim-O in more detail in Sections \ref{sec:rmfs} and \ref{rawsim}. After that we explain the four steps of the transition process from RMFS simulator to real-world RMFS shown in Figure \ref{fig:4_steps} in Sections \ref{sec:controller}, \ref{sec:erp}, \ref{sec:robots} and \ref{sec:robots_industry}, respectively. Section \ref{sec:conclusion} concludes our work and provides directions for future research.

\section{The Robotic Mobile Fulfillment System}\label{sec:rmfs}
\begin{figure}[h]
	\centering
	\includegraphics[width=\textwidth]{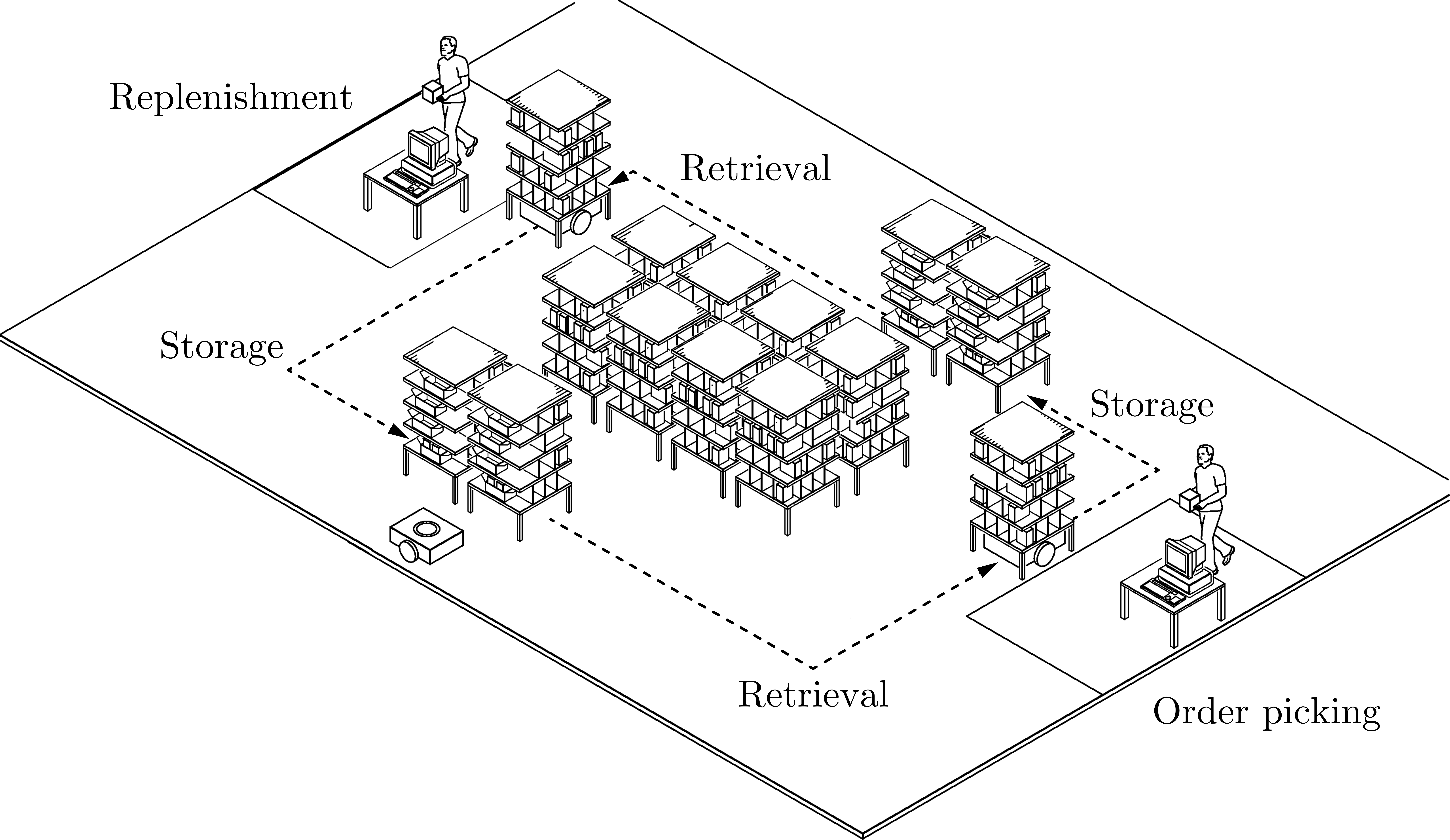}
	\caption{The central process of an RMFS (see \cite{Hoffman:2013}).}
	\label{fig:storage_retrieval_process}
\end{figure}

Instead of using a system of shelves and conveyors as in traditional parts-to-picker warehouses,
the central components of an RMFS are:
\begin{itemize}
	\item movable shelves, called \textit{pods}, on which the inventory is stored
	\item \textit{storage area} denoting the inventory area where the pods are stored
	\item workstations, where the pick order items are picked from pods (\textit{pick stations}) or replenishment orders are stored to pods (\textit{replenishment stations})
	\item mobile \textit{robots}, which can move underneath pods and carry them to workstations.
\end{itemize}
Firstly, we need to define some terms related to orders before explaining the processes in an RMFS, as follows:
\begin{itemize}
	\item stock keeping unit (\textit{SKU})
	\item an \textit{order line} including one SKU with number
	\item a \textit{pick order} including a set of order lines from a customer's order
	\item a \textit{replenishment order} consisting of a number of physical units of one SKU
\end{itemize}
The process of an RMFS is illustrated in Figure \ref{fig:storage_retrieval_process}. The pods are transported by robots between the storage area and workstations. Two processes are included:
\begin{itemize}
	\item \textit{retrieval process}: After the arrival of a replenishment order, robots are sent to carry the selected pods to a replenishment station, where the units are stored to these pods. We assume that a shared storage policy is applied (such as in \cite{bartholdi-hackman:2016}), which means SKUs of the same type are not stored together in a unique pod, but are spread over several pods.
	\item \textit{storage process}: After the arrival of a picking order, we get a set of order lines. Robots are selected to carry the required pods to a pick station, where the SKUs of the order lines are picked. We assume it is unlikely that a pick order can be completed with only one pod, unless there is only one order line or the association policy was applied in the retrieval process (in other words, all SKUs are stored together in one pod, if they are often ordered together by the same customer.)
\end{itemize}
 Then, after a pod has been processed at one or more stations, it is brought back to the storage area.

\subsection{Decision Problems}
In an RMFS environment, three levels of decision problems are required, namely the strategic, tactical and operational levels. The decision problems at the strategic level include storage area dimensioning and workstation placement (see \cite{lamballais2017estimating}), while the problems at the tactical level include decisions regarding the number of robots (see \cite{yuan2017bot}), and the numbers of pods and stations (see \cite{lamballais2017inventory}). Our simulation concentrates only on various operational decision problems for pods, robots and stations that have to be solved in an online e-commerce environment, including: 
\begin{itemize}
	\item \textit{Order Assignment} (orders to stations)
	\begin{itemize}
		\item \textbf{Replenishment Order Assignment (ROA)}: assignment of replenishment orders to replenishment stations
		\item \textbf{Pick Order Assignment (POA)}: assignment of pick orders to pick stations
	\end{itemize}
	\item \textit{Task Creation} (for pods)
	\begin{itemize}
		\item \textit{Pod Selection}
		\begin{itemize}
			\item \textbf{Replenishment Pod Selection (RPS)}: selection of the pod to store one replenishment order (see \cite{nigam2014analysis})
			\item \textbf{Pick Pod Selection (PPS)}: selection of the pods to use for picking the pick orders assigned at a pick station (see \cite{Boysen.2017} and \cite{zou2017assignment})
		\end{itemize}
		\item \textbf{Pod Repositioning (PR)}: assignment of an available storage location to a pod that needs to be brought back to the storage area (see \cite{merschformann2018active} and \cite{krenzler2018})
	\end{itemize}
	\item \textbf{Task Allocation (TA)} (for robots): assignment of tasks from \textit{Task Creation} and additional support tasks like idling to robots
	\item \textbf{Path Planning (PP)} (for robots): planning of the paths for the robots to execute (see \cite{Cohen.2015} and \cite{Cohen.2017})
\end{itemize}

\section{Simulation framework -- RAWSim-O}\label{rawsim}
\begin{figure}[htb]
	\centering
	\begin{subfigure}[b]{0.62\textwidth}
		\includegraphics[width = \textwidth]{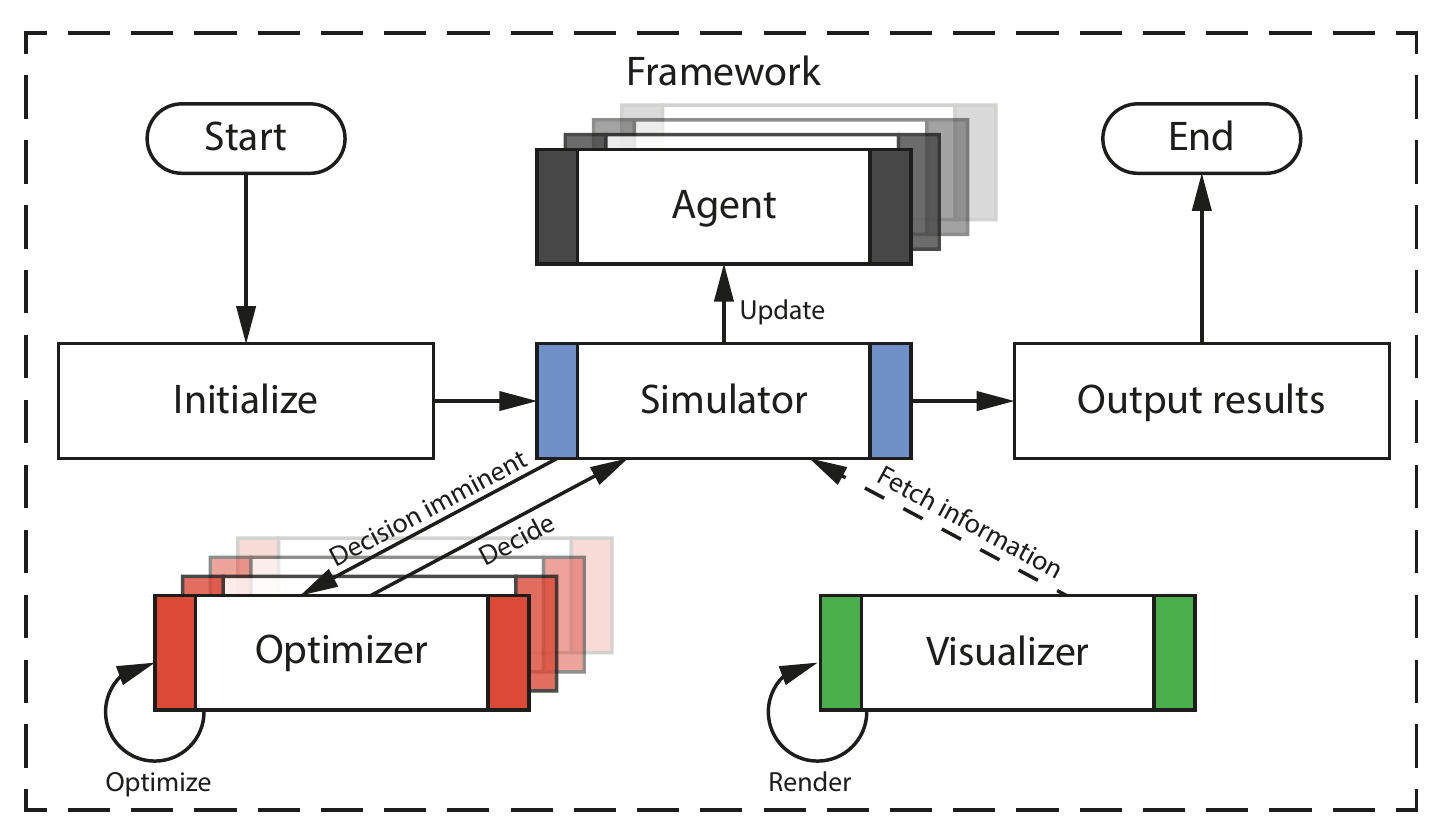}
		\caption{Overview of the simulation process.}
		\label{fig:mu_simulationframework}
	\end{subfigure}
	~
	\begin{subfigure}[b]{0.35\textwidth}
		\includegraphics[width = \textwidth]{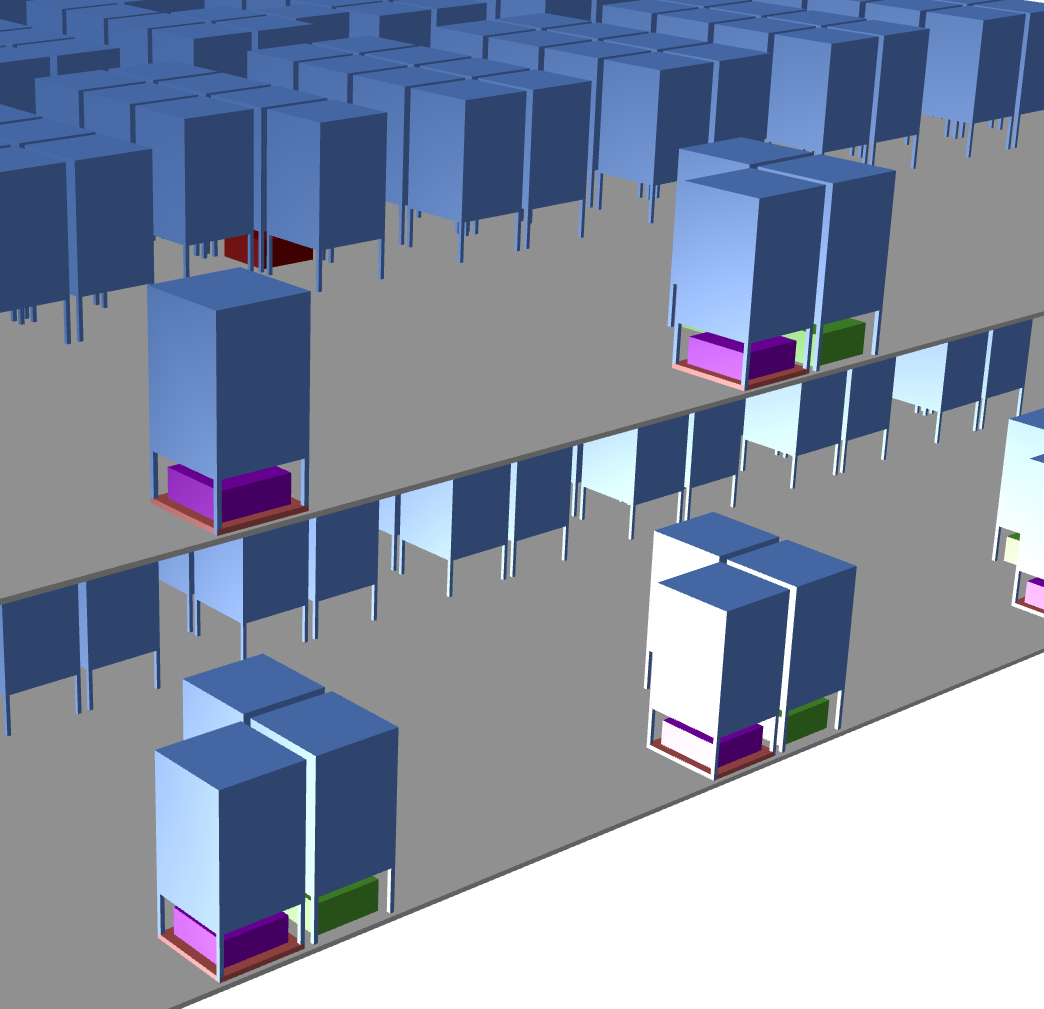}
		\caption{Visualization screenshot}
		\label{fig:mu_simulationscreenshot}
	\end{subfigure}
	\caption{RAWSim-O simulation framework.}
\end{figure}

RAWSim-O is an agent-based discrete-event simulation framework. It is designed to study the context of an RMFS while evaluating multiple decision problems jointly. 
Figure~\ref{fig:mu_simulationframework} shows an overview of our simulation process, which is managed by the core \textit{simulator} instance. The tasks of the simulator include:
\begin{itemize}
	\item Updating \textit{agents}, which can resemble either real entities, such as robots and stations, or virtual entities like managers, e.g. for emulating order processes.
	\item Passing decisions to \textit{optimizers}, which can either decide immediately or buffer multiple requests and release the decision later. 
	\item Exposing information to a \textit{visualizer}, which allows optional visual feedback in 2D or 3D. Figure~\ref{fig:mu_simulationscreenshot} illustrates a screenshot of our simulation in 3D.
\end{itemize}

\begin{figure}[t]
	\centering
	\includegraphics[width=0.85\textwidth]{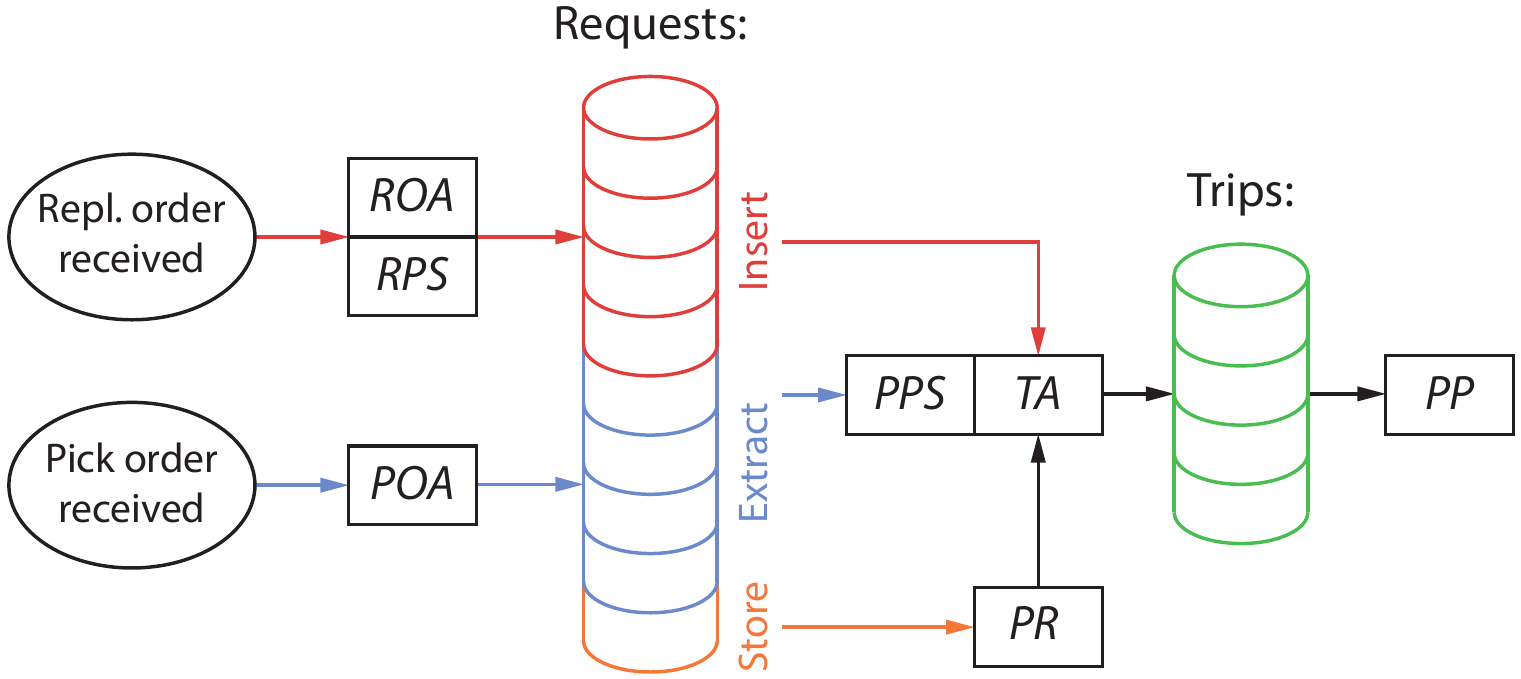}
	\caption{Order of decisions to be done triggered by receiving a pick or replenishment order.}
	\label{fig:problemdependencies}
\end{figure}

In the following, we describe the hierarchy of all core decision problems after new replenishment or pick orders are submitted to the system (see Figure~\ref{fig:problemdependencies}). If a new replenishment order is received, first the optimizers of ROA and RPS are responsible for choosing a replenishment station and a pod.
This technically results in an insertion request, i.e. a request for a robot to bring the selected pod to the given workstation.
A number of these requests are then combined in an insertion task and assigned to a robot by a TA optimizer.
Similarly, after the POA optimizer selects a pick order from the backlog and assigns it to a pick station, an extraction request is generated, i.e. a request to bring a suitable pod to the chosen station.
Up to this point, the physical units of SKUs for fulfilling the pick order are not yet chosen.
Instead, the decision is postponed and taken just before PPS combines different requests into extraction tasks and TA assigns these tasks to robots.
This allows the implemented optimizers to exploit more information when choosing a pod for picking. 
Hence, in this work we consider PPS as a decision closely interlinked with TA. 
Furthermore, the system generates store requests each time a pod is required to be transported to a storage location, and the PSA optimizer decides the storage location for that pod. The idle robots are located at dwelling points, which are located in the middle of the storage area to avoid blocking prominent storage locations next to the stations. Another possible type of task is charging, if the battery of a robot runs low; however, for this work we assume the battery capacity to be infinite. All of the tasks result in trips, which are planned by a PP algorithm. The only exception is when a pod can be used for another task at the same station, thus, not requiring the robot to move.

\begin{figure}[h]
	\centering
	\includegraphics[width=1\textwidth]{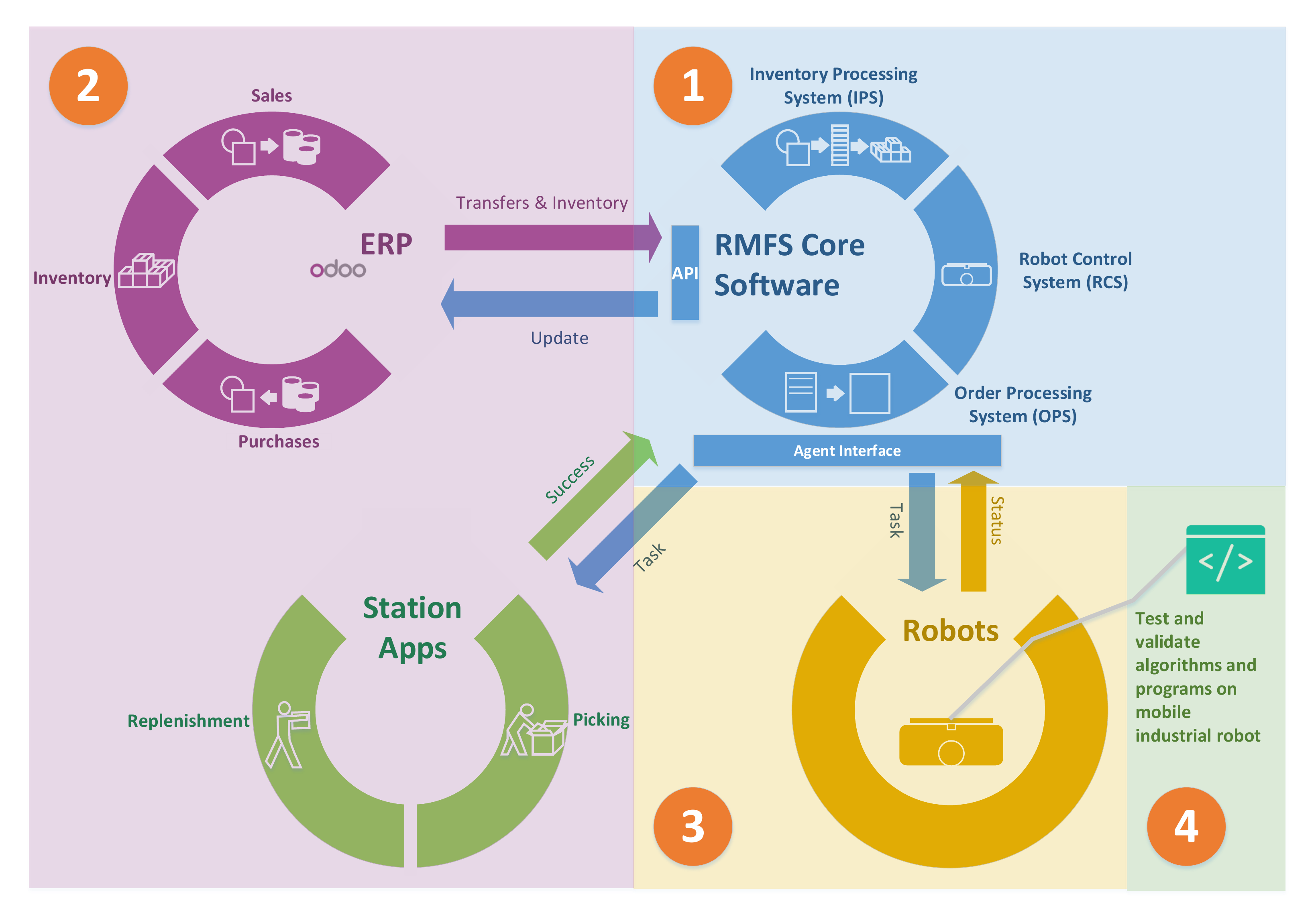}
	\caption{Relationships between the RMFS core software and the ERP system, the station apps, robots.}
	\label{fig:rmfs_4_steps}
\end{figure}

Figure~\ref{fig:rmfs_4_steps} gives an overview of an RMFS, which consists of an RMFS software system and robots. The simplest RMFS software system includes the RMFS core software, ERP and station apps. The numbers in Figure~\ref{fig:rmfs_4_steps} represent the four steps in the process for transforming, which is described in Figure \ref{fig:4_steps}. 

\section{Step 1: Changing RMFS simulator into the RMFS core software}\label{sec:controller}
The RMFS core software has three main parts, the inventory processing system (IPS), order processing system (OPS) and robot control system (RCS). The IPS functions as inventory management (the corresponding optimizers are RPS, PPS and PR), the OPS does transaction processing (the corresponding optimizers are ROA and POA), and the RCS operates and directs the robots (the corresponding optimizers are TA and PP). 
In this step, we extend the RMFS simulator to the RMFS core software. In total, there are two following extensions:

\begin{itemize}
	\item Implementation of application programming interface (API) for integrating with ERP and other enterprise systems.
	\item Implementation of the agent interface for robots and station apps.
\end{itemize}

The API of the RMFS core software is used for integrating with the ERP and other enterprise systems such as WMS (warehouse management system) and OMS (order management system). A simple description of the relationship between the ERP (or other enterprise systems) and the RMFS core software is that by using the API, the operational status and transactions will be reported back to the ERP, and the order information and the information of items stored in pods are passed from the ERP or other enterprise systems to the RMFS core software for processing. The API is designed for handling a variety of XML-RPC requests and XML-RPC responses (a remote procedure call (RPC) protocol \cite{laurent2001programming}), because the ERP system we used --
Odoo -- is easily available over XML-RPC with Odoo's external API. 

With an agent interface  (see  Figure~\ref{fig:rmfs_4_steps}) the core software can receive the status information from the robots. This includes the robot ID, the current robot position information, the orientation of the robot in radians, whether the pickup pod operation was successful and whether the setdown pod operation was successful etc. The core software can send task messages through the agent interface to the robots, which are listed in Table \ref{tab:taskmessage}, while it can receive some status messages from robots as shown in Table \ref{tab:statusmessage}. 

The agent interface is also used for exchanging various types of information between the core software and station apps. Table \ref{tab:station_infos} lists the important information for input/output stations, which is passed from the RMFS core software to the station apps for displaying to the replenisher or picker. The confirmation or error messages from the station apps are passed back to the RMFS core software to notify the operations in the input/output stations. 

\begin{table}
	\begin{center}
		\begin{tabular}{ r l}
			\textbf{Task message} & \textbf{Meaning}\\ 
			Go & execute move commands \\  
			Turn & execute turning commands for a defined number of \\
			&degrees to the right or left \\
			Rest & immediately stop all actions and rest where it is currently\\
			Pickup &  pick up a pod at the current position \\
			Setdown & set down a pod at the current position \\
			GetItem & its carrying pod is currently used for extraction \\
			PutItem & its carrying pod is currently used for insertion 
		\end{tabular}
		\caption{Meanings of task messages sent to robots.} \label{tab:taskmessage}
	\end{center}
\end{table}

\begin{table}
	\begin{center}
		\begin{tabularx}{\textwidth}{ r X}
			\textbf{Status message} & \textbf{Meaning}\\ 
			Error & explains an error that has occurred\\    
			WaypointTag & the current position of the robot\\
			Orientation & the orientation of the robot\\
			PickupSuccess & indicates whether the pickup operation was successful \\
			SetdownSuccess & indicates whether the setdown operation was successful \\
		\end{tabularx}
		\caption{Meanings of status messages received from robots.} \label{tab:statusmessage}
	\end{center}
\end{table}

\begin{table}
	\begin{center}
		\begin{tabularx}{\textwidth}{ r X}
			\textbf{Station information} & \textbf{Meaning}\\ 
			BundleID & ID of current bundle for replenishment (input station) \\    
			OrderID & ID of current order for picking (output station)\\
			ItemID & product ID (input/output station)\\
			Name & name of product/item (input/output station)\\
			Quantity & indicates quantity of product to put or get (input/output station)\\
			Pod Modeling Info & vertical and horizontal compartment placement within pod \\ 
			Stock Level Info & optimum stock level, maximum stock level and minimum presentation quantity (input station) \\
			Compartment Info & ID of product/item, filling rate or current count in this compartment (input/output station) \\
			Compartment to Pick & indicates the compartment in the pod to pick (output station)\\
			Compartment to Replenish & indicates the best compartment in the pod to replenish and the possible compartment to replenish for the current item (input station)\\			 
		\end{tabularx}
		\caption{Meaning of input/output station information received from the core software.} \label{tab:station_infos}
	\end{center}
\end{table}

\section{Step 2: Connecting the RMFS core software with ERP and station apps}\label{sec:erp}
\subsection{Connecting station apps}

In the automation of a real warehouse, the robot moves and carries the pods to the specified input/output stations. The pickers follow instructions on the output station app, grabbing items off the pod, while the replenishers follow instructions on the input station app and stuff products onto the pods for replenishing the inventory. Here, we explain how to connect station apps with our RMFS core software. 

\begin{figure}[h]
	\centering
	\begin{subfigure}[b]{0.45\textwidth}
		\includegraphics[width = \textwidth]{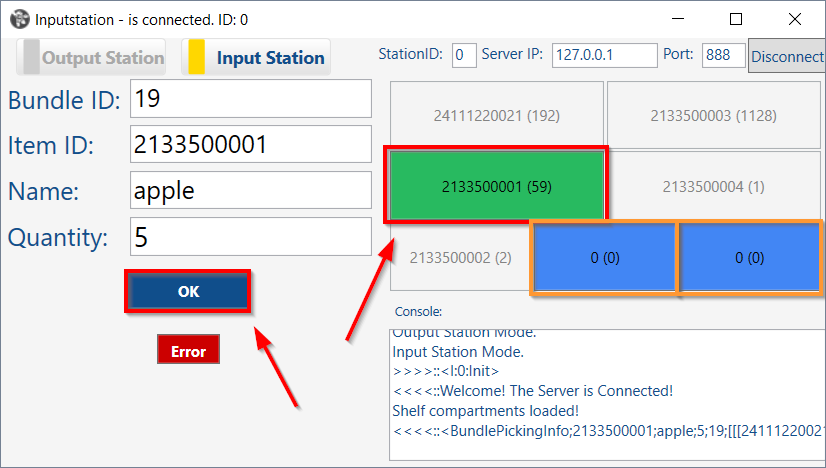}
		\caption{Input station GUI} \label{fig:inoutstation_gui_a}
	\end{subfigure}
	\begin{subfigure}[b]{0.45\textwidth}
		\includegraphics[width = \textwidth]{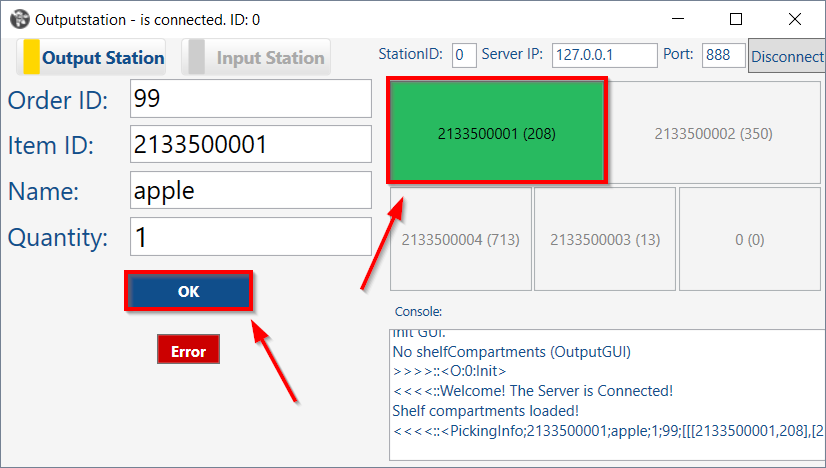}
		\caption{Output station GUI} \label{fig:inoutstation_gui_b}
	\end{subfigure}
	\caption{The GUI of the station apps.}
	\label{fig:inoutstation_gui}
\end{figure}

First, we develop graphical user interfaces in the programming language C\# that can be used at the input/output stations of the warehouse. To allow the use of several input and output stations in an automated warehouse, each station can be identified by its ID. 
Second, both applications connect to the agent interface via TCP and can receive, decode and display messages containing all the necessary information for the picker/replenisher. For example, as shown in the input station app for the replenishers in Figure~\ref{fig:inoutstation_gui_a}, the selected pod compartment is marked in green. Moreover, some other possible compartments are marked in blue. Similar to the picker, the selected pod compartment is marked as shown in Figure~\ref{fig:inoutstation_gui_b}. Pickers and replenishers can also send messages containing information about the success or failure of the picking/replenishment operation over the TCP connection by pressing the OK or Error Button.

\subsection{Connecting Odoo}

Additionally, the API of IPS and the OPS in the RMFS core software are defined and implemented to exchange information with Odoo's external API (\url{https://odoo.com/documentation/11.0/webservices/odoo.html}), which can send and receive information using XML-RPC. We mainly use three parts of Odoo, namely purchases, sales and inventory. We use the Inventory module of Odoo to manage the location and contents of all pods and all kinds of transfers (item movement from a to b), whether incoming, outgoing or internal. We also use the Sales and Purchase modules to create sales orders and purchase orders. 

\subsubsection{Order to Picking}
\begin{figure}[h]
	\centering
	\begin{subfigure}[b]{0.45\textwidth}
		\includegraphics[width = \textwidth]{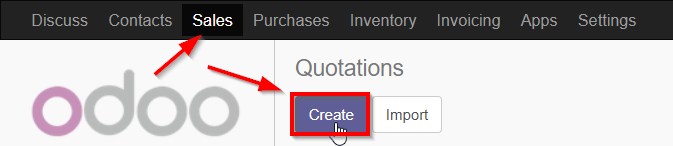}
		\caption{Create a sale order} \label{fig:example_a}
		\includegraphics[width = \textwidth]{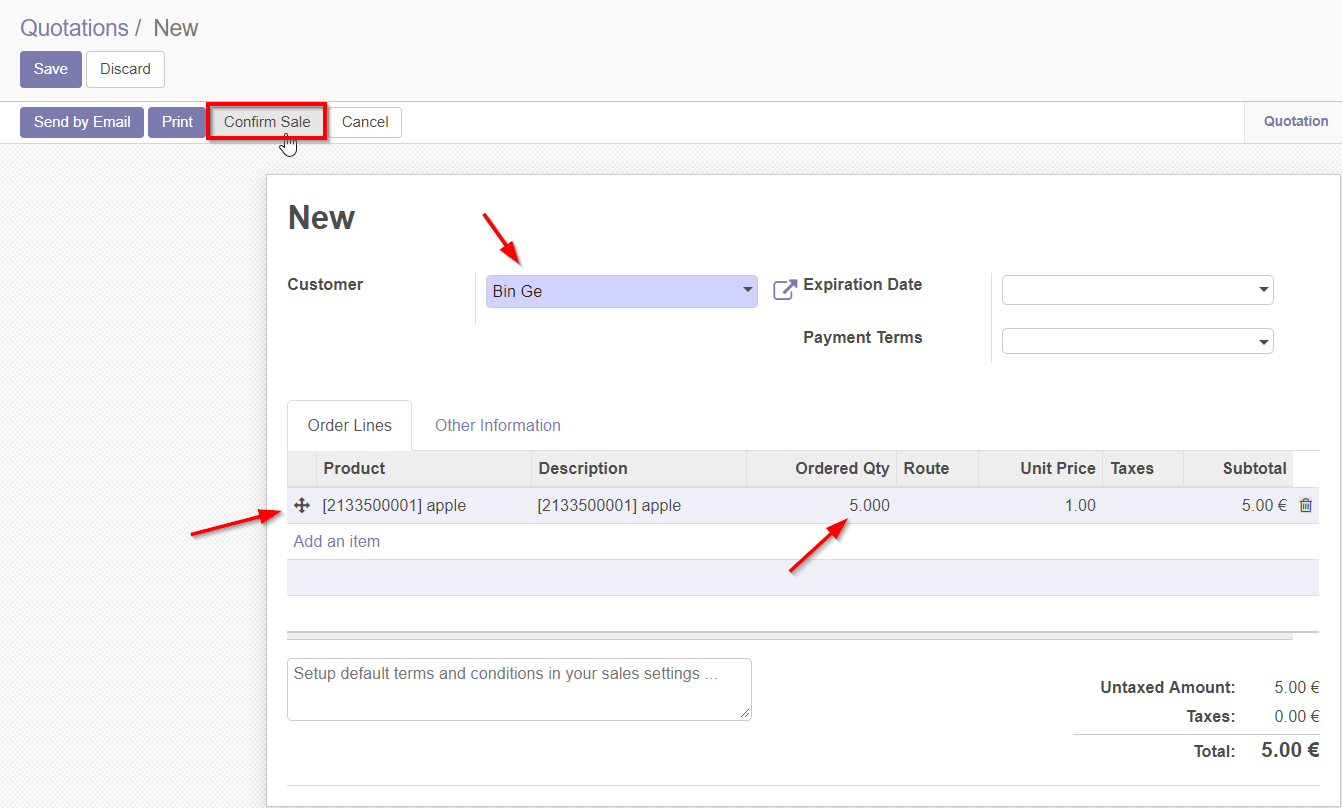}
		\caption{Sale information} \label{fig:example_b}
	\end{subfigure}
	\begin{subfigure}[b]{0.45\textwidth}
		\includegraphics[width = 0.3\textwidth]{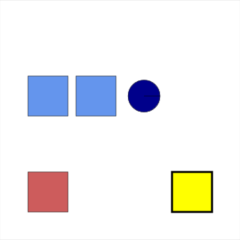}
		\includegraphics[width = 0.3\textwidth]{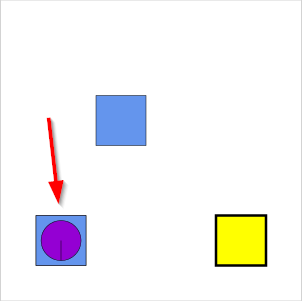}
		\includegraphics[width = 0.3\textwidth]{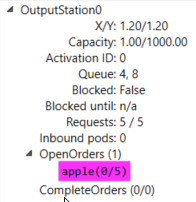}
		\caption{RAWSim-O view} \label{fig:example_c}
		\includegraphics[width = 0.85\textwidth]{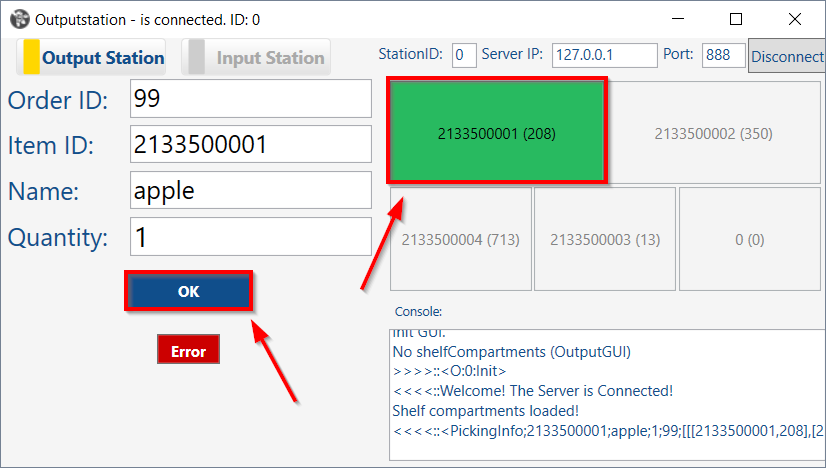}
		\caption{Picking station} \label{fig:example_d}
	\end{subfigure}
	\caption{An example from generating a sale order in Odoo to picking.}
	\label{fig:process_example}
\end{figure}

In this part we will describe the process of an order passing through the system, from receiving an order in Odoo to the package leaving the warehouse. We describe this process firstly with a small example shown in Figure~\ref{fig:process_example}. After an order is created with a demand for five apples (Figure~\ref{fig:example_a}) in Odoo, the sale is confirmed for sale in Figure~\ref{fig:example_b}. The RMFS core software then recognizes the incoming order and selects a pod (or several, if needed) containing at least five apples and the pod is moved by a robot to the designated picking station (Figure~\ref{fig:example_c}). The picker then taks the items from the pod and confirm the sucessful picking operations by pressing the OK buttom in the output station app (Figure~\ref{fig:example_d}). The green marked position is the pod compartment where the items are located in the pod. The output station app gets this information from a JSON (JavaScript Object Notation)-formatted message from the RMFS core software agent interface. After a successful picking operation, the core software adjusts the inventory of the pod that was used.

\begin{figure}[h]
	\centering
	\includegraphics[width=\textwidth]{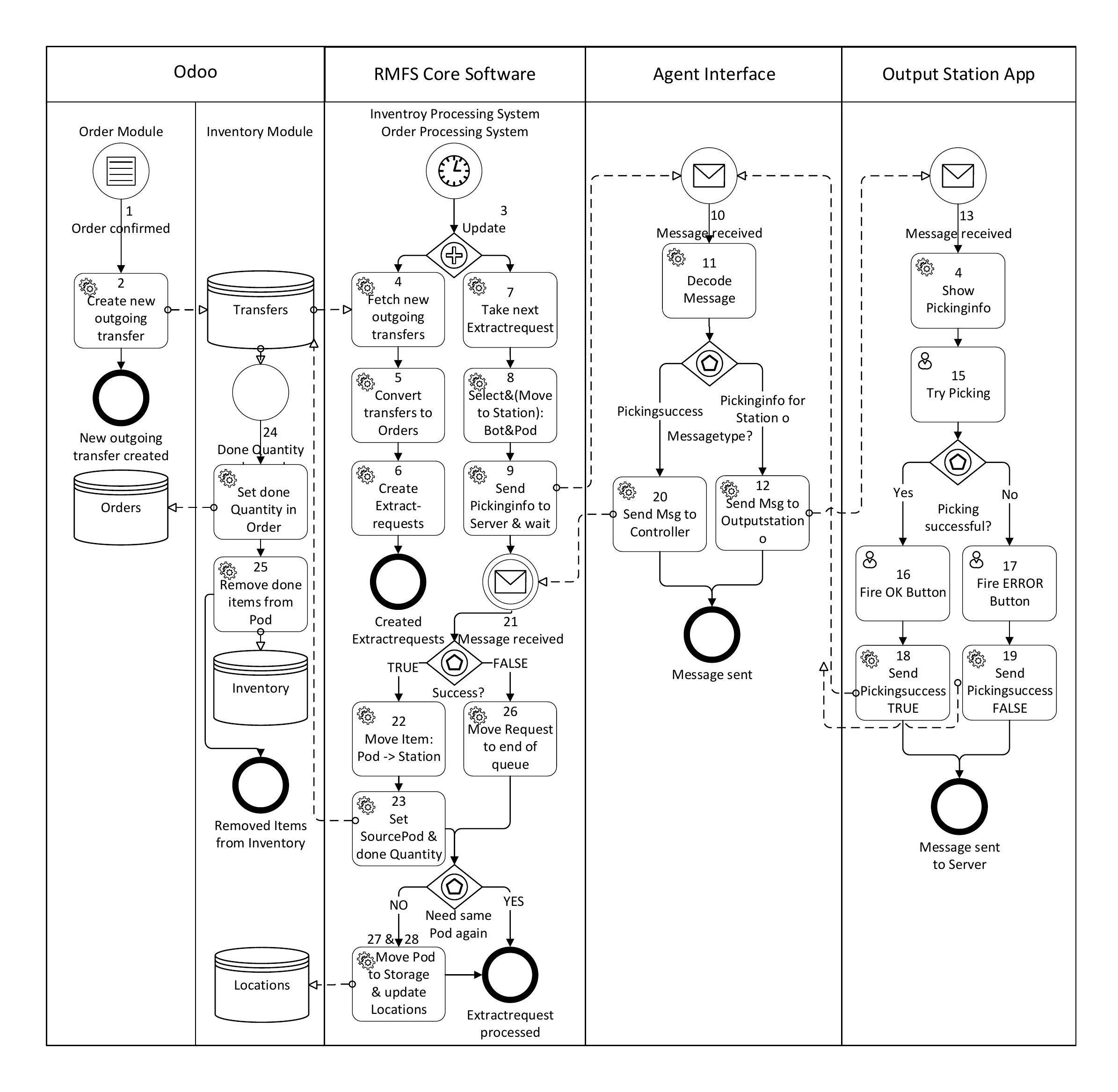}
	\caption{The process diagramm of an order passing through the system.}
	\label{fig:process}
\end{figure}

We will now explain more about the technical details, referring to Figure \ref{fig:process}. We will use some Odoo-specific terms, and some descriptions in the picking process, such as enqueuing and dequeuing of extract requests, are simplified. On startup, the RMFS core software retrieves the current positions and contents of all pods from Odoo.
When an order is placed (1), Odoo creates a new planned, outgoing transfer, from Stock to Customer (2).  Whenever the Update method in the RMFS core software is called (3), it scans the Odoo Transfer database for new planned, outgoing transfers (4). The transfers are converted to orders (5), for which the RMFS core software then generates and enqueues extract requests (6). An extract request is a request to pick a specific item belonging to a certain order at a chosen output station. Concurrently, the RMFS core software dequeues an extract request (7) whenever the Update method is called (3), chooses a pod to fulfill the request and moves the pod to the designated output station (8), unless it is already there because it was used to fulfill the previous extract request. Once the pod arrives at the output station, the RMFS core software sends a message containing the ID of the output station and information about the item and the affiliated order over a previously established TCP/IP connection to the agent interface (9). Once the interface receives a message (10), it decodes the message (11) and identifies its type. As in this case the type is “Pickinginfo”, it then forwards the item's name and ID and the order ID to the output station that was specified in the message (12). Once the output station app receives the message (13), it displays the received information (14) and waits for the picker to finish the picking operation (15) and to respond with “OK” if it was successful (16) or “Error” if the item could not be picked (17). The information about the success of the picking operation is then sent to the agent interface (18/19). The agent interface forwards the output station's response and ID to the RMFS core software (10$\rightarrow$11$\rightarrow$20). If the output station's response (21) is an “OK”, the RMFS core software moves the item from the pod to the station in the visualization/simulation (22), sets the pod as the move's source location in Odoo and increments the done quantity of the move by 1 (23). If this completes the last move of a transfer, the transfer is marked as done in Odoo. Odoo reacts to an increase of the done quantity (24) by removing the same quantity of the item from the source location (25). If the output station has sent an error message, the RMFS core software skips the extract request and moves it to the end of the queue (26). Finally, the RMFS core software checks if it can use the same pod again to fulfill the next extract request. If the same pod cannot be used again, a storage location will be chosen and the pod will be moved (27). Once the robot sets the pod down, the RMFS core software updates the pod's location in Odoo (28). 

\subsubsection{Purchase Order to Replenishment}
The replenishment process's implementation is rather similar to the picking process described above, therefore only the main differences will be described. The event triggering the start of the replenishment process in the RMFS core software is the receipt of the replenishment items. After that, an internal transfer is automatically created to transfer the received items from the receiving area into pods. When this transfer is detected by the RMFS core software, this transfer is split into its moves and they are converted to item bundles (amount x of sku s). So an insert request for each item bundle is generated (as opposed to an extract request for each order line), which is then handled by the RMFS core software.

\section{Step 3: Integrating the RMFS core software with mini-robots in XOR-bench}\label{sec:robots}
Teleworkbench \cite{tanoto2009teleworkbench} and  \cite{tanoto2011teleworkbench} is a platform or an infrastructure for conducting, analyzing and evaluating experiments using a number of mini-robots. It offers a controlled environment in which users can execute and test the robot programs using real robots \cite{tanoto2011teleworkbench}. Similar to Teleworkbench, we designed a teleoperated platform (call XOR-bench), but its features are specially designed for an RMFS. The XOR-bench includes the following functionalities:

\begin{itemize}
	\item Live video of experiment 	
	\item Program-download to robot
	\item Computer vision-based robot positioning system 
	\item Events and messages logger
	\item Wireless communication system
	\item Internet connectivity			
\end{itemize}

The real-time experiment video will be streamed through an IP-based camera. The user can download the developed robot program to the robots through XOR-bench. The robots' positions will be obtained by using computer vision technology from the captured live video. All events and messages from the RMFS software system (including the RMFS core software, ERP and station apps) are recorded and accessible. With the logger data the researchers and developers can analyze the results of the experiments. The XOR-bench provides wireless communication between the robots and the RMFS core software. The XOR-bench is connected to the Internet to enable easy access for researchers and developers.

\begin{figure}[h]
	\centering
	\includegraphics[width=\textwidth]{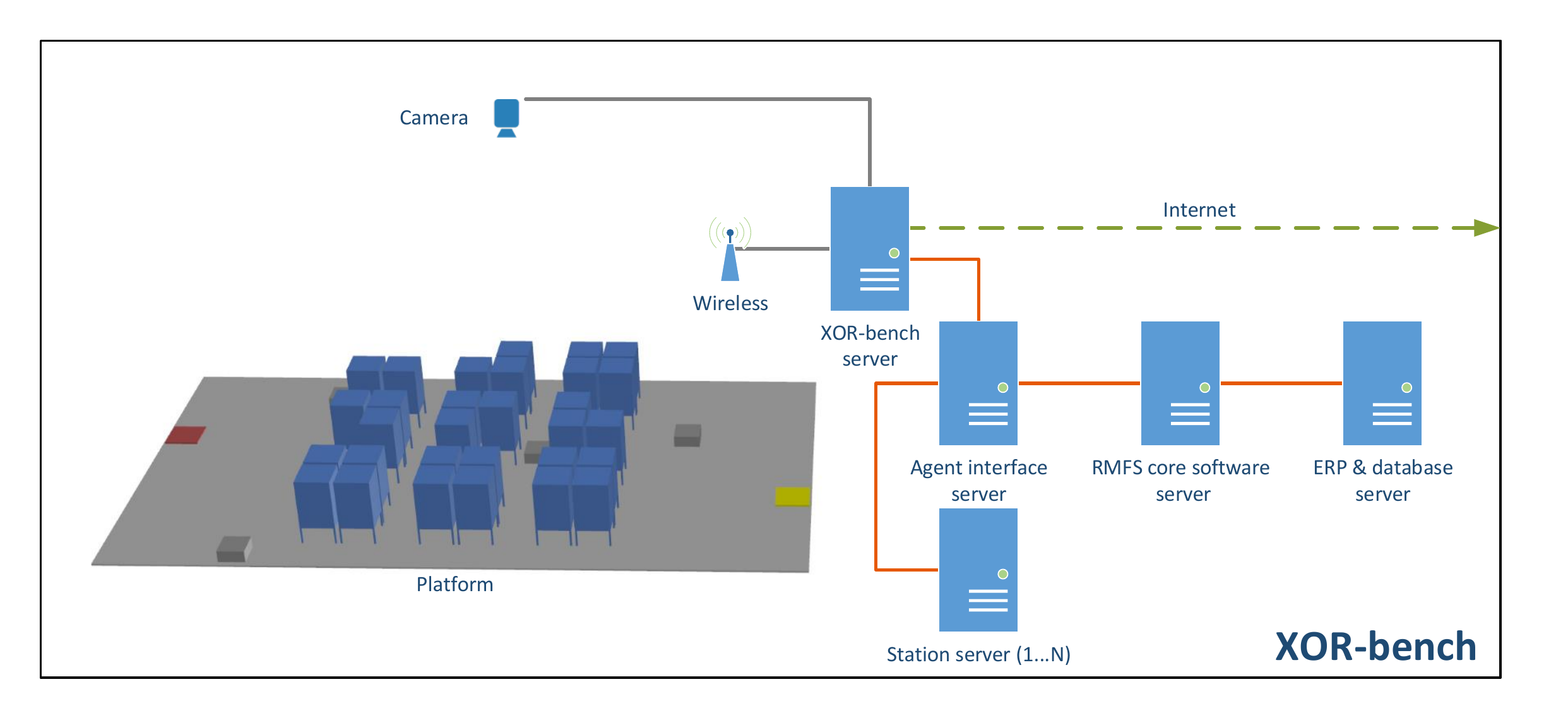}
	\caption{The general system architecture of the XOR-bench system.}
	\label{fig:xor-bench_architecture}
\end{figure}

Figure~\ref{fig:xor-bench_architecture} illustrates the general system architecture of the XOR-bench system. The XOR-bench server provides a TCP/IP service, which can be used to connect the robots (the gray rectangles in the platform) with the agent interface sever. The agent interface server is connected with the RMFS core software server and the station servers. The station apps and ERP (including the database) are located in the station servers and ERP \& database server respectively. Each station app is located in a station server.

In this step, we use mini-robots, because they are affordable and support high programmability and expandability in the research and teaching. Unlike using mobile industrial robots, using mini-robots for the RMFS system integration experiment, there are some significant advantages, for instance, the experiment is easy to set up, execute, monitor and analyze in a short time. Several RMFS system integration experiments can be done rapidly.

Two different mini robotic platforms are currently used on the XOR-bench. In our previous research work \cite{Merschformann-xie-li:2017}, we used the iRobot Create 2, a mobile robot platform based on the Roomba vacuum cleaning robot. The robots are equipped with ASUS Eee PCs through serial-to-USB cables for processing capabilities, and web-cams for line-following. The RFID tag reader is mounted inside the former vacuum cleaning compartment for waypoint recognition. The robot program was written in C\#. The main drawback of using iRobot Create 2 for the RMFS experiment is that the robot does not have the lifting mechanism to elevate the pods off the floor when transporting them.

\begin{figure}[h]
	\centering
	\begin{subfigure}[b]{0.45\textwidth}
		\includegraphics[width = \textwidth]{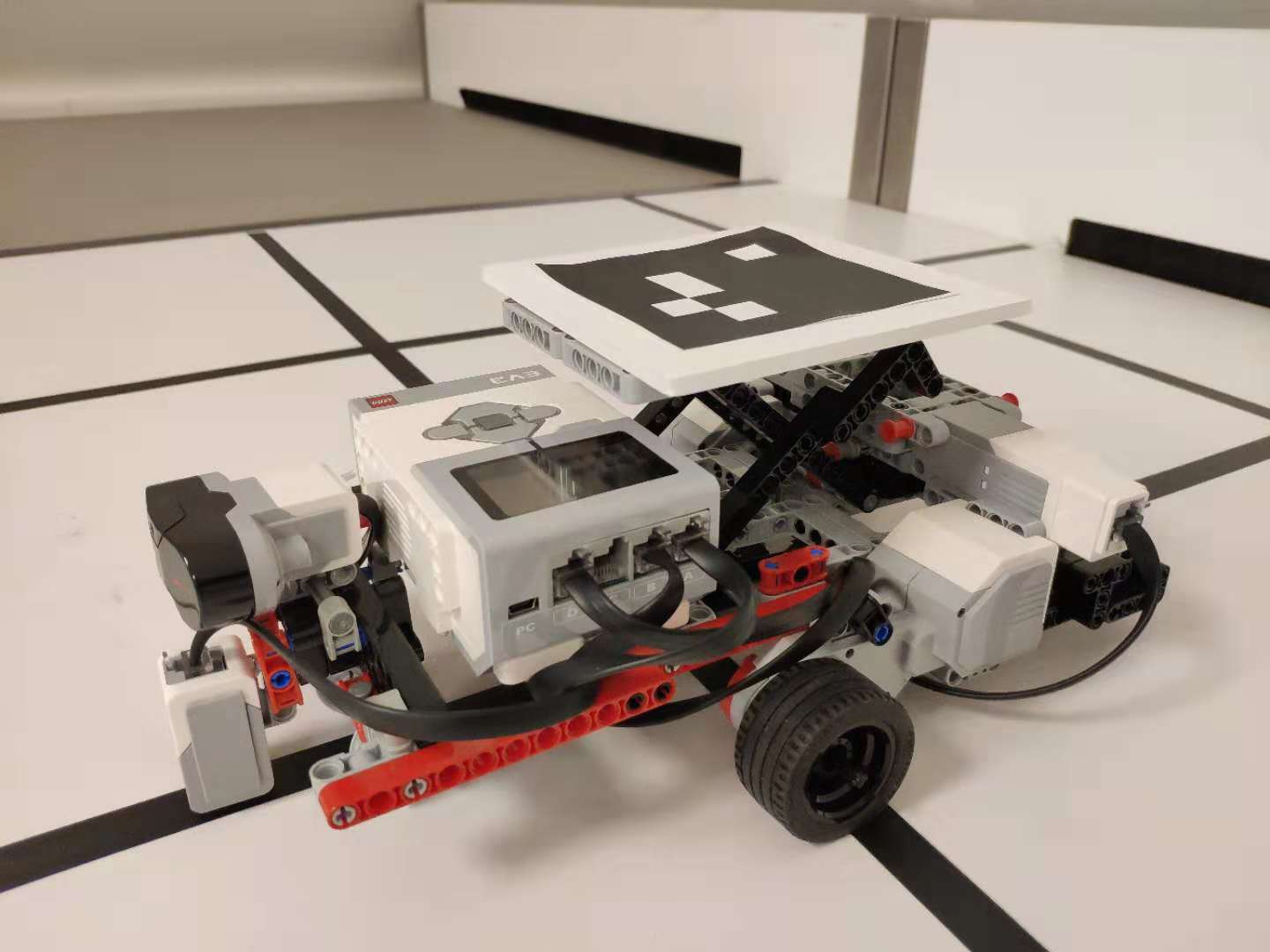}
		\caption{Close-up of the robot} \label{fig:lego_rmfs_robot_a}
	\end{subfigure}
	\begin{subfigure}[b]{0.45\textwidth}
		\includegraphics[width = \textwidth]{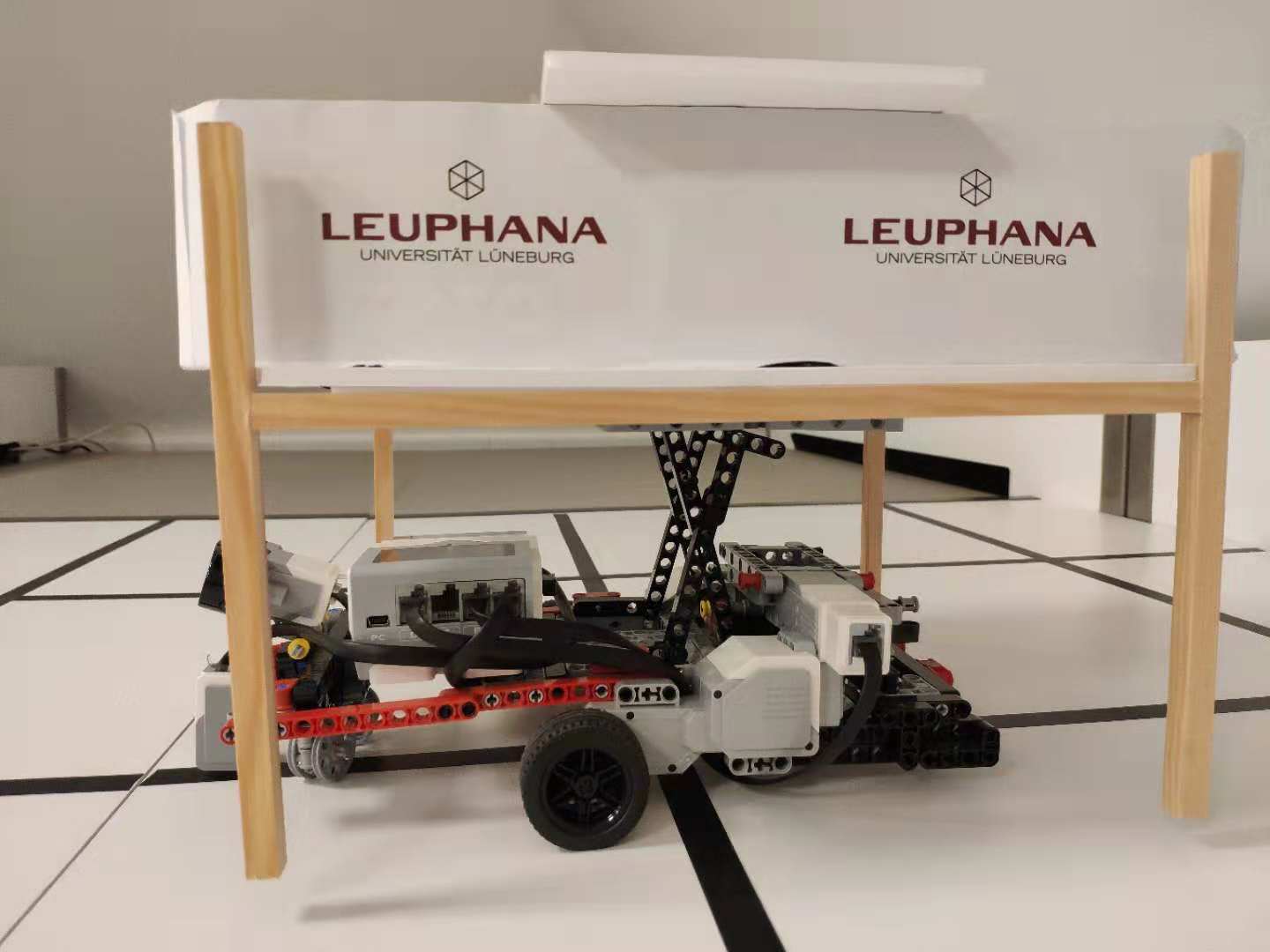}
	\caption{Robot carries the pod} \label{fig:lego_rmfs_robot_b}
	\end{subfigure}
	\caption{LEGO-RMFS robot.}
	\label{fig:lego_rmfs_robot}
\end{figure}

The new mini-robot (see Figure \ref{fig:lego_rmfs_robot_a}) we use on the XOR-bench is built by LEGO Mindstorms EV3 (third-generation robotics kit in LEGO's Mindstorms line). The robotics kit makes it easy to build the robot, with ease of programming, low complexity and low costs both for the research and teaching. The robot has three motors, with one middle motor used for the lifting mechanical structure (the robot can pick up or set down the pod: see Figure \ref{fig:lego_rmfs_robot_b}), while two large motors are differential motors for the movement. We also program the robot similar to typical RMFS robots, such as line-following, turning right and turning left.  A color sensor is responsible for the function of line-following; moreover, an ultrasonic sensor is used for avoiding obstacle and distance detection. The program was written in the EV3 Python programming language to control the robot, which runs on the EV3dev operating system (a modified version of the Linux Debian Jessie operating system). 

We set the maximum velocity limit of each robot to 0.05 m/s, while the time it takes for each robot to do a complete turn is set to 3 s. The time for the robot to set down and pick up the pod is about 3 s. The commands are sent to the robot via WiFi. 

\begin{figure}[h]
	\centering
	\begin{subfigure}[b]{0.45\textwidth}
		\includegraphics[width = \textwidth]{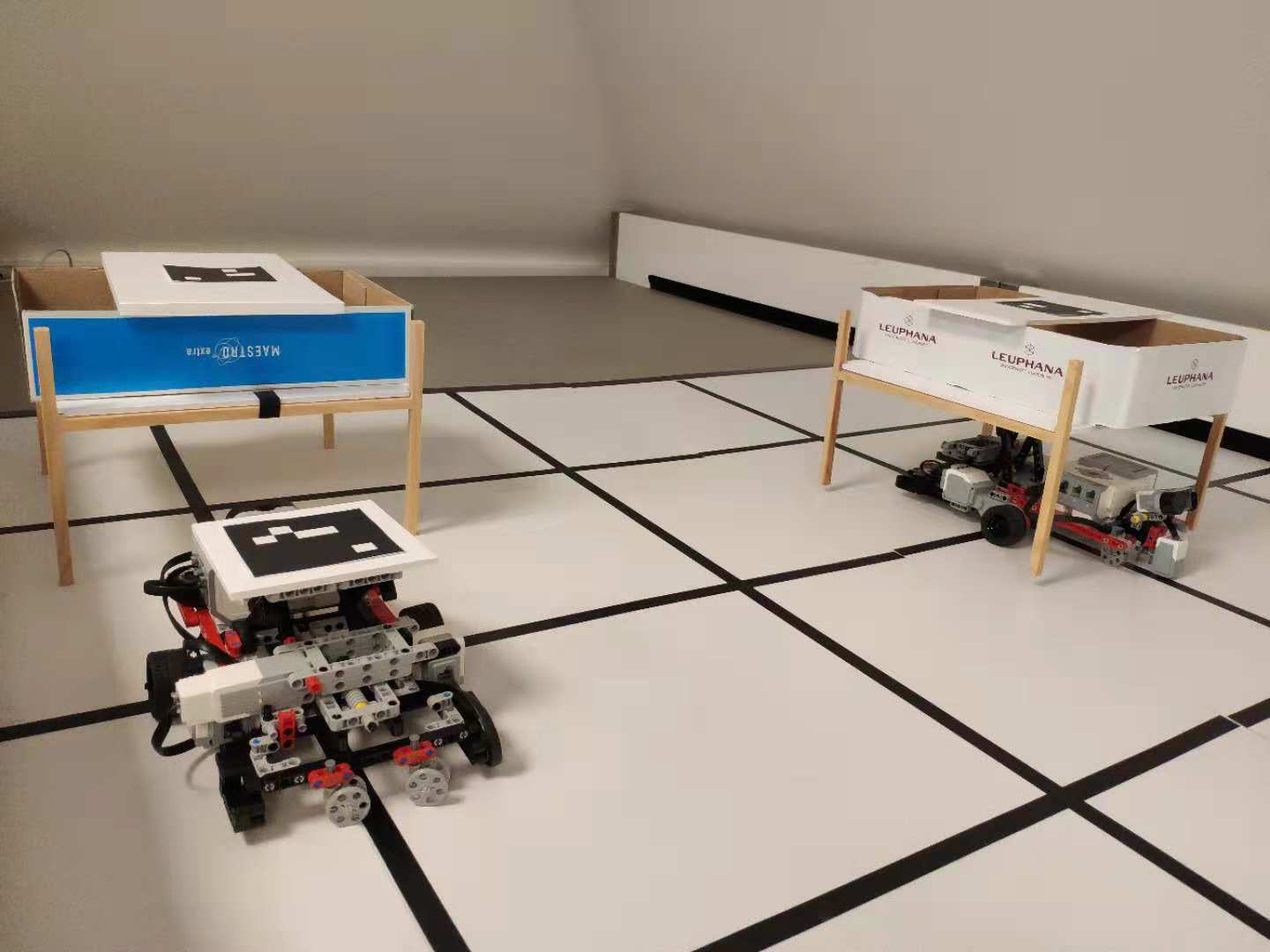}
		\caption{Robots in the experiment} \label{fig:lego_exp_a}
	\end{subfigure}
	\begin{subfigure}[b]{0.45\textwidth}
		\includegraphics[width = \textwidth]{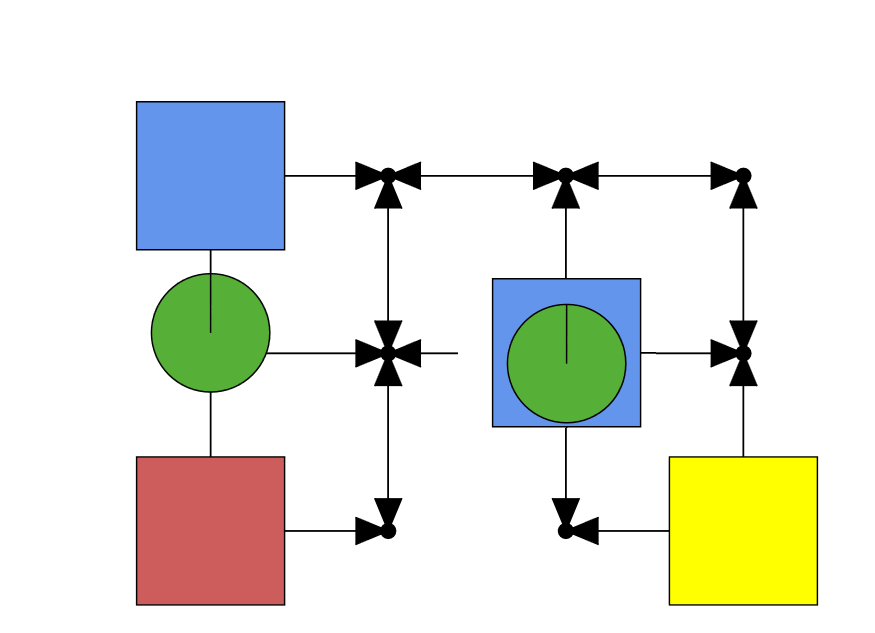}
		\caption{Visualization in RMFS core software} \label{fig:lego_exp_b}
	\end{subfigure}
	\caption{Experiment with LEGO-RMFS robots.}
	\label{fig:lego_rmfs_experiment}
\end{figure}

In the experiment as shown in Figure~\ref{fig:lego_rmfs_experiment}, we use a  $3\times4$ grid layout. One replenishment station is set in the bottom-left corner and one pick station is set on the bottom-right corner. Two robots are used; the pods on the field are made with straight sticks. Figure~\ref{fig:lego_exp_b} shows the visualization of the experiment in our RMFS core software. The robot is in the green circle, which carries the pod in the blue rectangle, while the red rectangle is the output station and the yellow one is the input station. 

\begin{figure}[h]
	\centering
	\begin{subfigure}[b]{0.45\textwidth}
		\includegraphics[width = \textwidth]{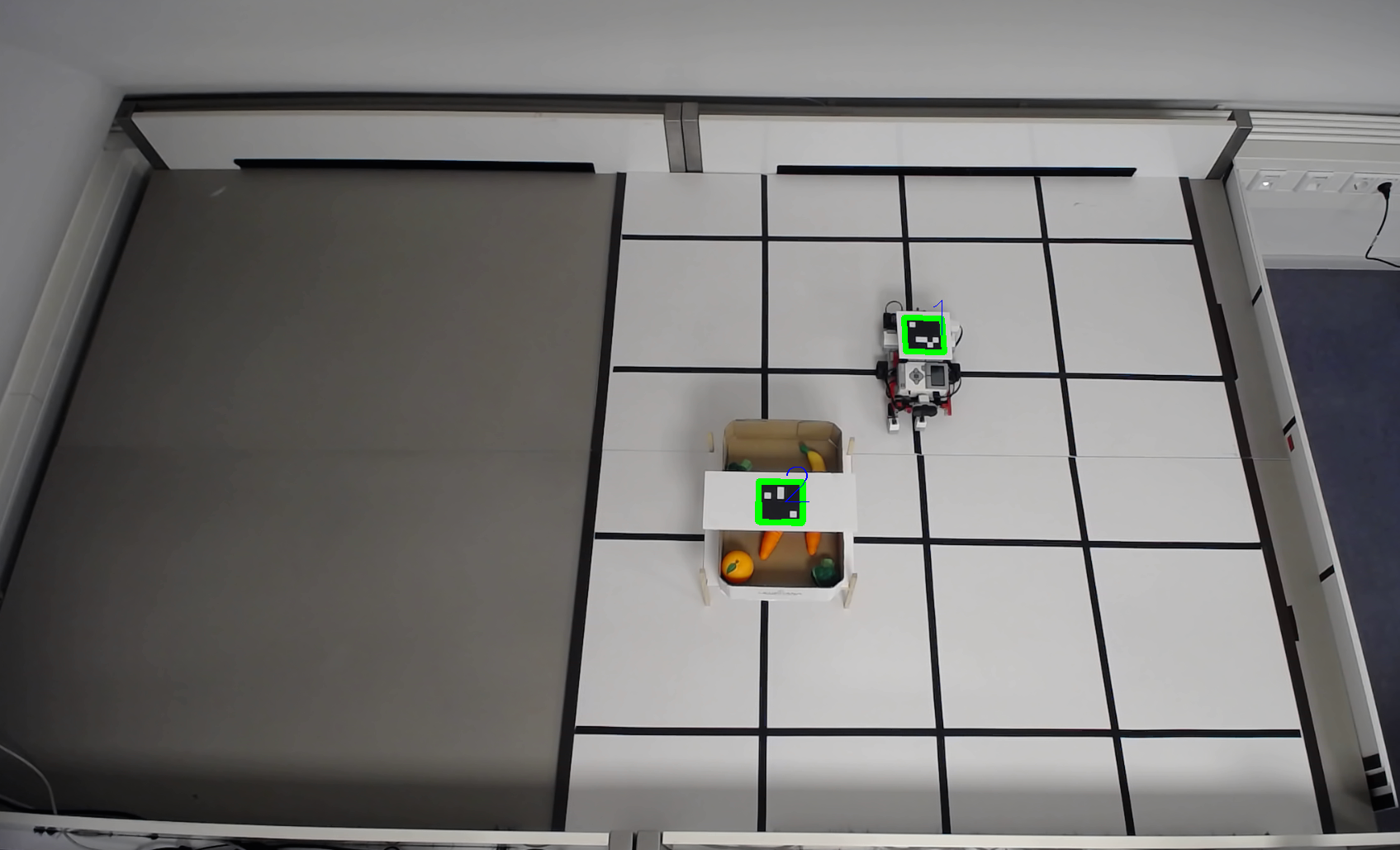}
		\caption{Robot positioning in the experiment platform} \label{fig:lego_visual_a}
	\end{subfigure}
	\begin{subfigure}[b]{0.2\textwidth}
		\includegraphics[width = \textwidth]{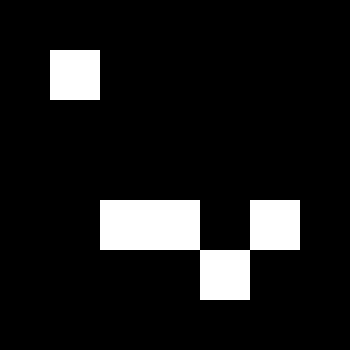}
		\caption{Example of ArUco marker} \label{fig:lego_visual_b}
	\end{subfigure}
	\caption{LEGO-RMFS robot positioning by using of computer vision technology.}
	\label{fig:rmfs_robot_positioning}
\end{figure}

As shown in Figure \ref{fig:lego_visual_a}, the XOR-bench in the lab has a computer vision-based multi-robot positioning system similar to that in \cite{tanoto2012scalable}, so the robot position information as well as identification of different robots can be captured. The software components of the robot positioning are written in Python with OpenCV (Open Source Computer Vision Library). On each pod and robot, an ArUco marker in Figure~\ref{fig:lego_visual_b} is used for localization. We use an open-source algorithm for detecting the ArUco marker as in \cite{garrido2014automatic}. The robot positioning software runs on XOR-bench server and sends the robot position information to the RMFS core software.

\section{Step 4: Test and validate algorithms and programs on a mobile industrial robot in XOR-bench}\label{sec:robots_industry}
In the last step of the process from RMFS simulation to the real world, we perform the experiment in the XOR-bench for testing and validating the on-board algorithms and programs of a mobile industrial robot. The mobile industrial robots are expensive; moreover, they need a large place to execute the experiments. Therefore, the test and validation of algorithms and programs on them is highly complex. Ideally using the Internet connectivity feature of the XOR-bench, we can remotely perform experiments for industrial robots by using real-time information streamed over the Internet, such as video, and exchanged messages between the robots and the RMFS core software. We can do the robot tele-programming (in other words, we can develop a robot program locally and it is possible to download it to remote robots).

\begin{figure}[h]
	\centering
	\begin{subfigure}[b]{0.4\textwidth}
		\includegraphics[width = \textwidth]{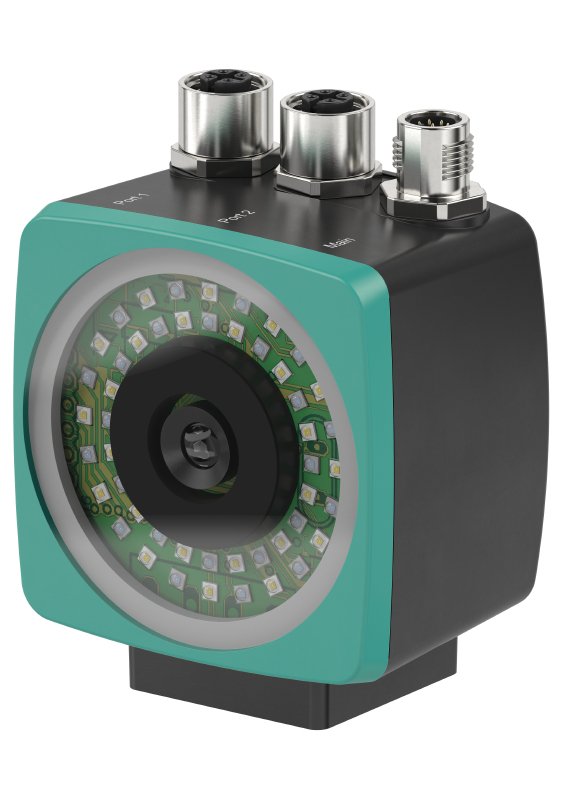}
		\caption{Pepperl+Fuchs PGV position tracking and control system} \label{fig:real_robot_positioning_a}
	\end{subfigure}
	\begin{subfigure}[b]{0.4\textwidth}
		\includegraphics[width = \textwidth]{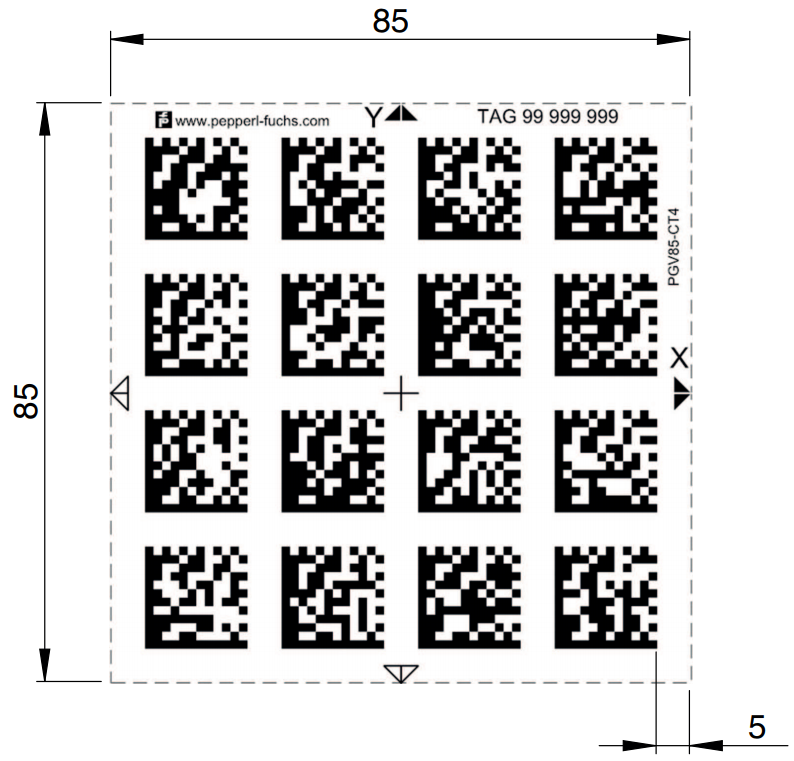}
		\caption{Data Matrix tag with the number 99999999 and position information} \label{fig:real_robot_positioning_b}
	\end{subfigure}
	\caption{Real robot positioning by using of position tracking and control system.}
	\label{fig:real_robot_positioning}
\end{figure}

The mobile industrial robot for the remote experiment is the Xiellog-Z series from Hanning ZN Tech Beijing, China. The payload of the robot is 650 kg; the speed of the robot is 1.2 m/s. The robot measures about 830mm $\times$ 650mm $\times$ 350mm and about 150 kg. The robot mounts an advanced industrial positioning tracking and control system on the base, which can be used for data matrix-based code positioning tracking and guiding the robot along a colored path. In other words, the system looks down at the ground to recognize matrix-based codes and color lines on the floor. Figure~\ref{fig:real_robot_positioning_a} shows a data matrix-based position tracking and control system called PGV from the German sensor manufacturing company Pepperl+Fuchs. Using an industrial 2-D camera, the PGV can also guides a robot along its colored path. Figure~\ref{fig:real_robot_positioning_b} shows a data matrix tag containing position information in addition to a specific number (\url{https://www.pepperl-fuchs.com/global/en/classid_3334.htm}). Because of the robot's on-board positioning system, for this remote experiment, we do not need the video-based positioning system in the XOR-bench. 

The robot has a scissor-lift mechanism to lift and pods. Moreover, they have an industrial WiFi module, which can be connected with the application server through the TCP/IP protocol. The controller on the robot is a Lenze C300 PLC (Programmable Logic Controller), which requires high reliability control. The program on the robot was written in high-level-language-structured text under the IEC 61131-3 standard.

\begin{figure}
	\centering
	\includegraphics[width=\textwidth]{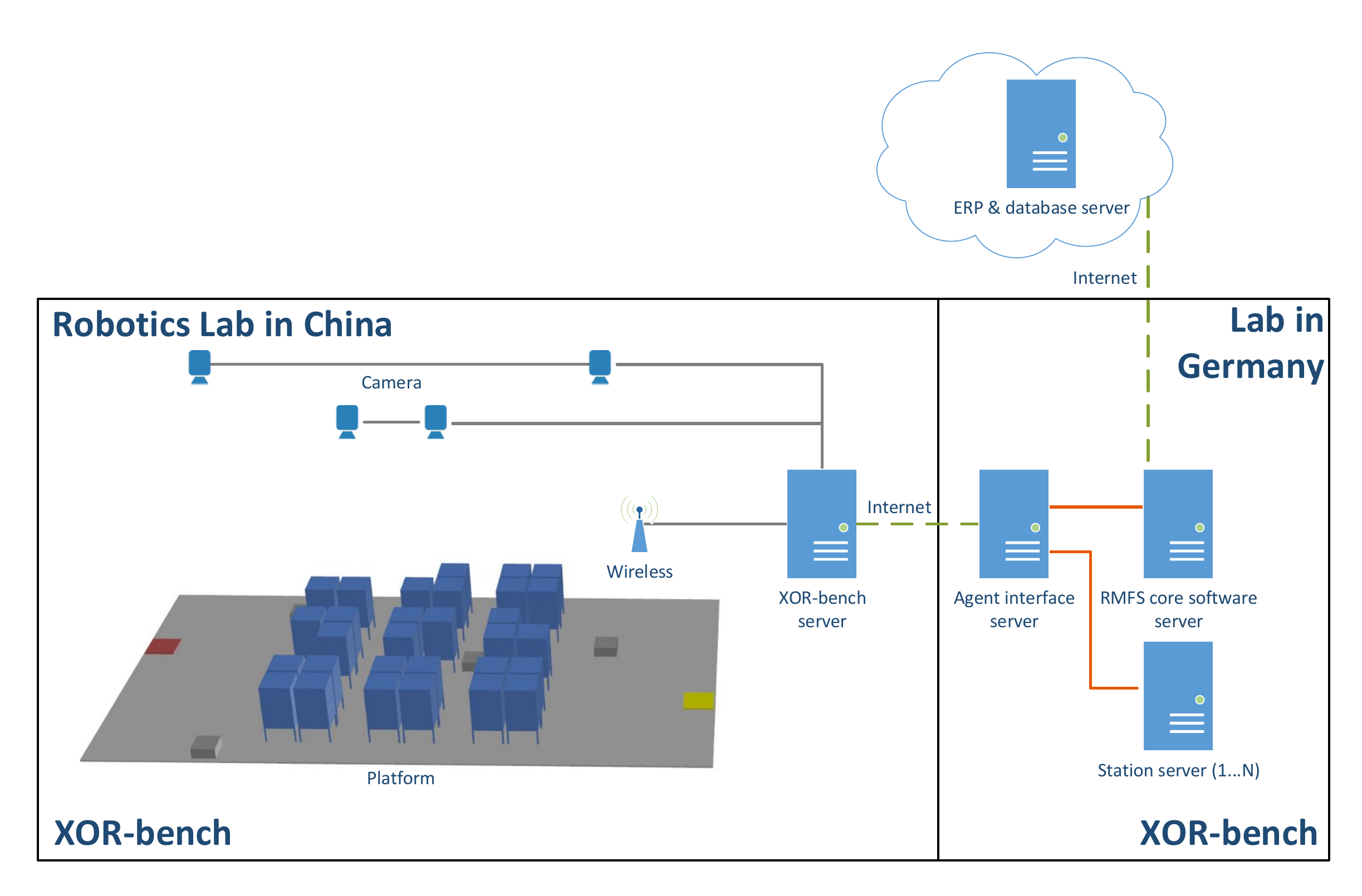}
	\caption{The system architecture of the XOR-bench system in remote testing situation.}
	\label{fig:xor-bench_remote}
\end{figure}

The infrastructure we designed allows people around the world to remotely test and benchmark the robot programs, without spending a lot of effort on handling communication and experiment settings. Figure~\ref{fig:xor-bench_remote} shows the system architecture of the XOR-bench system in the remote test situation. The XOR-bench is split into two parts: one part located in the robotics lab in Beijing, China, which includes the XOR-bench server, four IP-based cameras and mobile industrial robots; another part located in our lab in Germany, which has the RMFS core software server agent interface server and the station server. Note that the station server is in the real world located with the station; however, it is in our case located in the lab in Germany for the experiment. The Odoo ERP system and databases are hosted on the Tencent cloud. 

\begin{figure}
	\centering
	\begin{subfigure}[b]{0.4\textwidth}
		\includegraphics[width = \textwidth]{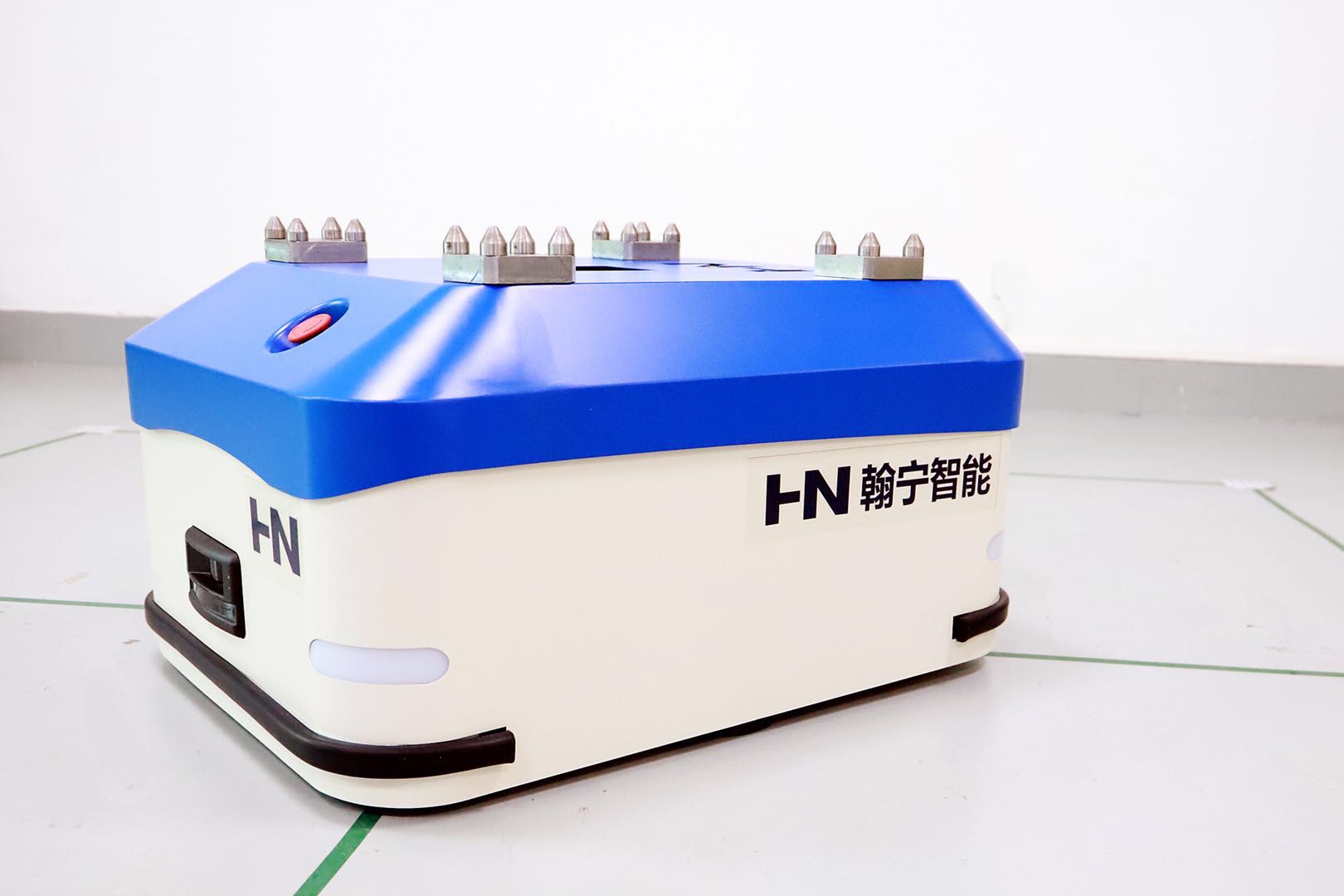}
		\caption{Close-up of the robot} \label{fig:real_robot_experiment_a}
		\includegraphics[width = \textwidth]{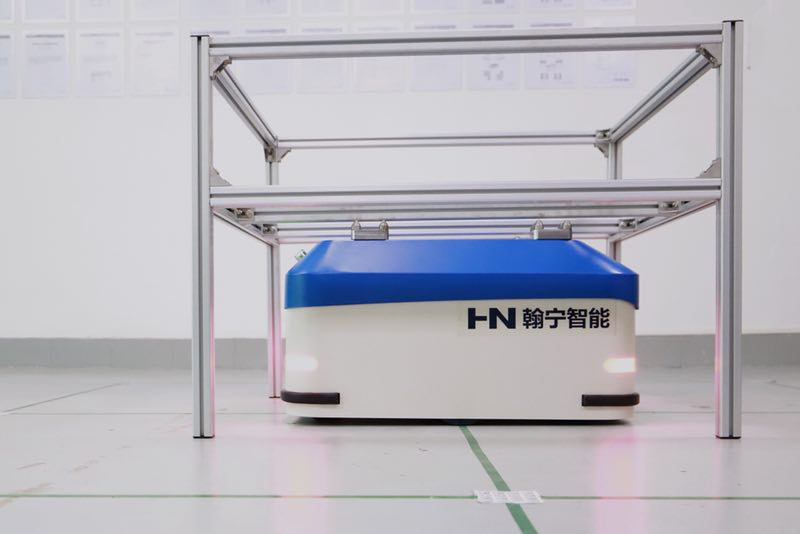}
		\caption{Robot carries the pod} \label{fig:real_robot_experiment_b}	
	\end{subfigure}
	\begin{subfigure}[b]{0.4\textwidth}
		\includegraphics[width = \textwidth]{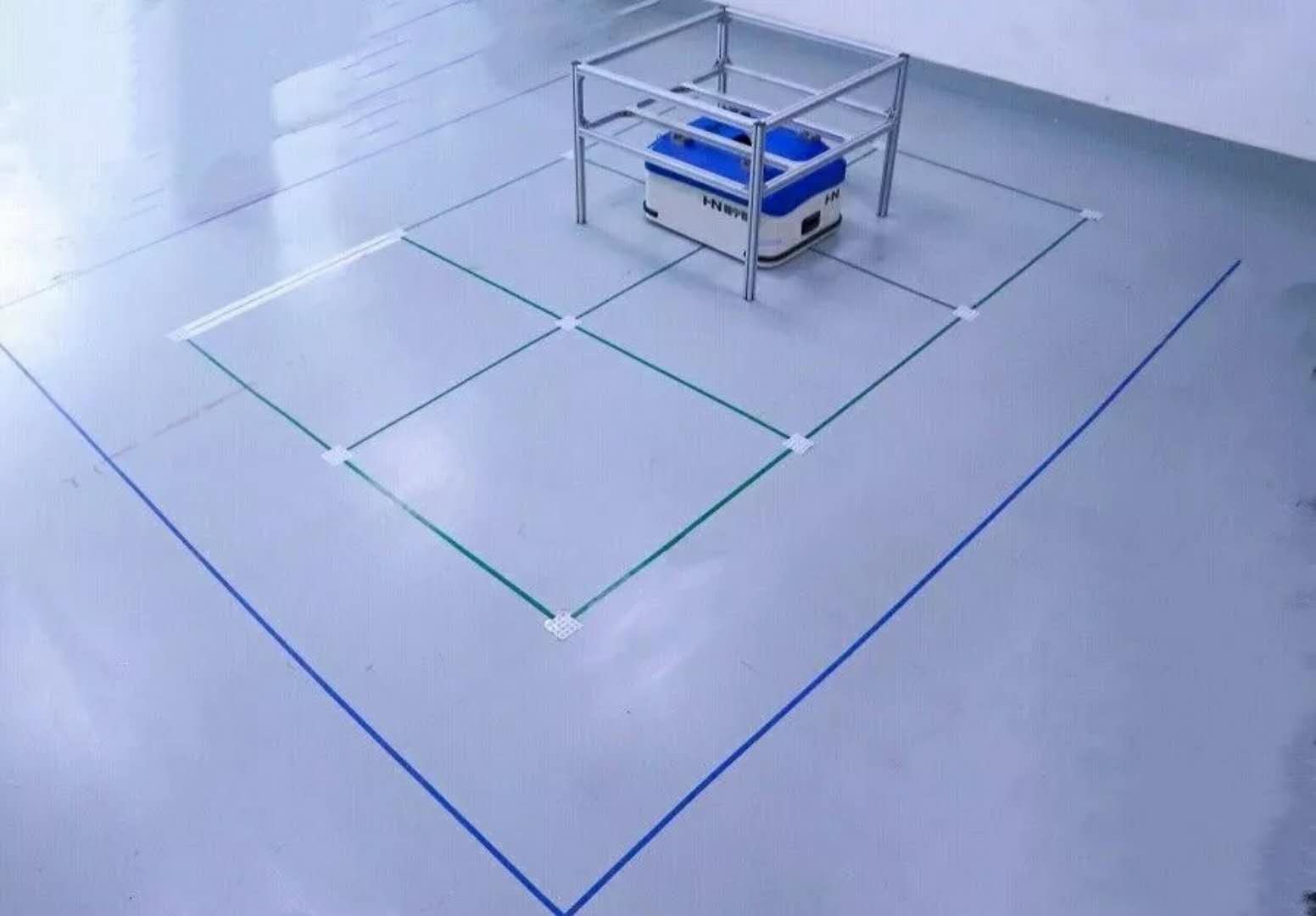}
		\caption{Robot in the experiment} \label{fig:real_robot_experiment_c}
		\includegraphics[width = \textwidth]{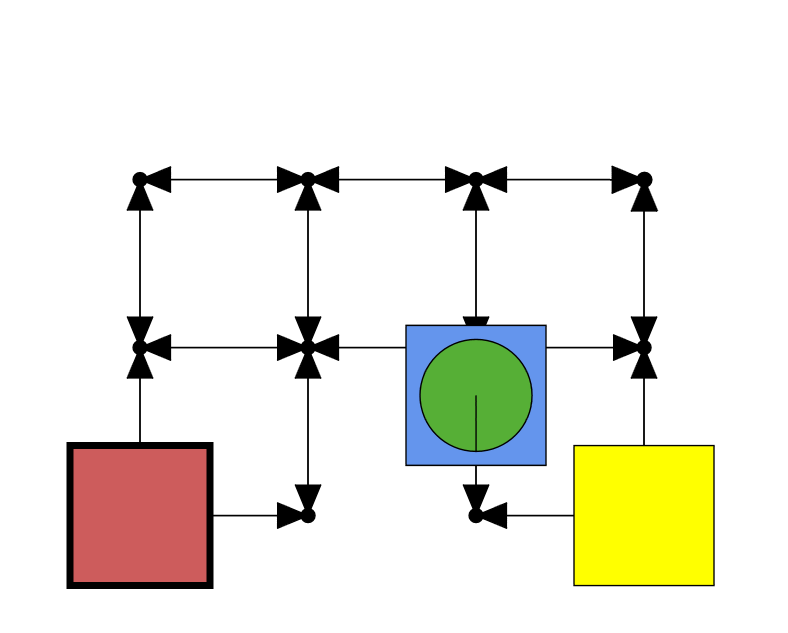}
		\caption{Visualization in RMFS core software} \label{fig:real_robot_experiment_d}
	\end{subfigure}
	\caption{Remote experiment with mobile industrial robot.}
	\label{fig:real_robot_experiment}
\end{figure}

As with the experiment layout in the previous section, a $3\times4$ grid layout is used for performing the experiment with a mobile industrial robot. Figures~\ref{fig:real_robot_experiment_a}~\ref{fig:real_robot_experiment_b} show the mobile industrial robot. Figure~\ref{fig:real_robot_experiment_c} gives a view of the IP-based camera in the lab during the experiments, while Figure~\ref{fig:real_robot_experiment_d} shows the visualization of the information in the RMFS core software. Recall that the robot is in the green circle, which carries the pod in the blue rectangle, while the red rectangle is the output station and the yellow one is the input station. 

\section{Summary and outlook}\label{sec:conclusion}
In order to answer the question at the beginning of this paper: “how can we make sure that the algorithms we implement and test in the simulator are still appliable for the real-world scenarios?”, we give you an example of an optimizer we developed in the simulation, the path planning optimizer. And we will explain how we make this optimizer appliable for the real-world scenarios. The path planning algorithm is called Multi-Agent Pathfinding (MAPF) in the literature. This is a challenging problem with many applications such as robotics, disaster rescue and video games (see \cite{Wang.2011}). The goal in MAPF is to compute collision-free paths along a grid-based graph for all agents from their start to their unique destination nodes. There are many MAPF solvers from artificial intelligence which typically work for agents, but they do not consider kinematic constraints, such as maximum velocity limits, maximum acceleration and deceleration, and turning times. Moreover, they consider the movement of robots only in discretized environments. It is a similar case with the existing path planning algorithms for an RMFS (see \cite{Cohen.2015} and \cite{Cohen.2017}). However, we can not use those MAPF solvers directly if we want to test the path planning algorithms with real robots. Moreover, there are MAPF solvers from robotics, which consider the kinematic constraints, but they work only for a small number of robots; therefore, they are too slow for a massive search like in an RMFS. In order to make path planning algorithms work for real robots, we have implemented new MAPF solvers in our path planning optimizer that consider kinematic constraints in a continuous environment for the RMFS. More details can be found in \cite{merschformann2017path}.

In order to make it easy to test the developed algorithms in the simulator with real-world data and robots, we have presented in this work a novel development process flow by using our XOR-bench. This process flow includes changing the RMFS simulator into core software and integrating the RMFS core software with Odoo ERP and input/output station app. 

Moreover, we have shown that geographically distributed users can use our designed XOR-bench to test and validate the RMFS system and algorithms on real robots, either educational robots (such as iRobots or LEGO robots) or industrial robots. This is beneficial for both research and teaching.

Due to the existing interfaces with the ERP system and the robots, we can use the XOR-bench in the future to test other types of automated warehousing systems, such as an automated system \cite{ackerman_2016}, where humans work in zones, and robots are sent to humans to load items from zones to input/output stations; in this case, robots work alongside humans (traveling in the same space with humans).

\section{Acknowledgements}
We would like to thank Beijing Hanning ZN Tech for providing the robots and robotics research lab and performing the RMFS system experiment with us. 

\bibliographystyle{plain}
\bibliography{erp-arxiv}

\begin{thebibliography}{10}

\bibitem{ackerman_2016}
Evan Ackerman.
\newblock How locus robotics plans to build a successor to {Amazon's Kiva}
  robots, Aug 2016.

\bibitem{bartholdi-hackman:2016}
J.~J. Bartholdi and S.~T. Hackmann.
\newblock Warehouse \& distribution science.
\newblock Supply Chain and Logistics Institute. Release 0.98, 2017.

\bibitem{Boysen.2017}
Nils Boysen, Dirk Briskorn, and Simon Emde.
\newblock Parts-to-picker based order processing in a rack-moving mobile robots
  environment.
\newblock {\em European Journal of Operational Research}, 262(2):550--562,
  2017.

\bibitem{Cohen.2017}
L.~Cohen, G.~Wagner, T.~K. {Satish Kumar}, H.~Choset, and S.~Koenig.
\newblock Rapid randomized restarts for multi-agent path finding solvers.
\newblock {\em ArXiv e-prints}, 2017.

\bibitem{Cohen.2015}
Liron Cohen, Tansel Uras, and Sven Koenig.
\newblock Feasibility study: Using highways for bounded-suboptimal multi-agent
  path finding.
\newblock In {\em Eighth Annual Symposium on Combinatorial Search}, 2015.

\bibitem{de-koster-le-duc-roodbergen:2018}
René de~Koster, Tho Le-Duc, and Kees~Jan Roodbergen.
\newblock Design and control of warehouse order picking: A literature review.
\newblock {\em European Journal of Operational Research}, 182(2):481--501,
  2007.

\bibitem{garrido2014automatic}
Sergio Garrido-Jurado, Rafael Mu{\~n}oz-Salinas, Francisco~Jos{\'e}
  Madrid-Cuevas, and Manuel~Jes{\'u}s Mar{\'\i}n-Jim{\'e}nez.
\newblock Automatic generation and detection of highly reliable fiducial
  markers under occlusion.
\newblock {\em Pattern Recognition}, 47(6):2280--2292, 2014.

\bibitem{Hazard.2006}
Christopher~J. Hazard, Peter~R. Wurman, and Raffaello D'Andrea.
\newblock {Alphabet Soup}: A testbed for studying resource allocation in
  multi-vehicle systems.
\newblock In {\em Proceedings of AAAI Workshop on Auction Mechanisms for Robot
  Coordination}, pages 23--30. Citeseer, 2006.

\bibitem{Hoffman:2013}
A.~E. Hoffman, M.~C. Mountz, M.~T. Barbehenn, J.~R. Allard, M.~E. Kimmel,
  F.~Santini, M.~H. Decker, R.~D'Andrea, and P.~R. Wurman.
\newblock System and method for inventory management using mobile drive units,
  2013.

\bibitem{krenzler2018}
Ruslan Krenzler, Lin Xie, and Hanyi Li.
\newblock Deteministic pod repositioning problem in robotic mobile fulfillment
  systems.
\newblock 2018.

\bibitem{lamballais2017estimating}
Tim Lamballais, Debjit Roy, and MBM De~Koster.
\newblock Estimating performance in a robotic mobile fulfillment system.
\newblock {\em European Journal of Operational Research}, 256(3):976--990,
  2017.

\bibitem{lamballais2017inventory}
Tim Lamballais, Debjit Roy, and MBM De~Koster.
\newblock Inventory allocation in robotic mobile fulfillment systems.
\newblock 2017.

\bibitem{laurent2001programming}
Simon~St Laurent, Joe Johnston, Edd Wilder-James, and Dave Winer.
\newblock {\em Programming Web Services with XML-RPC: Creating Web Application
  Gateways}.
\newblock O'Reilly Media, Inc., 2001.

\bibitem{Merschformann-xie-li:2017}
M.~Merschformann, L.~Xie, and H.~Li.
\newblock {RAWSim-O}: A simulation framework for robotic mobile fulfillment
  systems.
\newblock {\em Logistics Research}, 11(1), 2018.

\bibitem{merschformann2018active}
Marius Merschformann.
\newblock Active repositioning of storage units in robotic mobile fulfillment
  systems.
\newblock In {\em Operations Research Proceedings 2017}, pages 379--385.
  Springer, 2018.

\bibitem{merschformann2017path}
Marius Merschformann, Lin Xie, and Daniel Erdmann.
\newblock Path planning for robotic mobile fulfillment systems.
\newblock {\em arXiv preprint arXiv:1706.09347}.

\bibitem{nigam2014analysis}
Shobhit Nigam, Debjit Roy, Rene de~Koster, and Ivo Adan.
\newblock Analysis of class-based storage strategies for the mobile shelf-based
  order pick system.
\newblock 2014.

\bibitem{tanoto2012scalable}
Andry Tanoto, Hanyi Li, Ulrich R{\"u}ckert, and Joaquin Sitte.
\newblock Scalable and flexible vision-based multi-robot tracking system.
\newblock In {\em 2012 IEEE International Symposium on Intelligent Control
  (ISIC)}, pages 19--24. IEEE, 2012.

\bibitem{tanoto2009teleworkbench}
Andry Tanoto, Ulrich R{\"u}ckert, and Ulf Witkowski.
\newblock Teleworkbench: A teleoperated platform for experiments in
  multi-robotics.
\newblock In {\em Web-Based Control and Robotics Education}, pages 267--296.
  Springer, 2009.

\bibitem{tanoto2011teleworkbench}
Andry Tanoto, Felix Werner, Ulrich R{\"u}ckert, and Hanyi Li.
\newblock Teleworkbench: validating robot programs from simulation to
  prototyping with minirobots.
\newblock In {\em The 10th International Conference on Autonomous Agents and
  Multiagent Systems}, volume~3, pages 1303--1304. International Foundation for
  Autonomous Agents and Multiagent Systems, 2011.

\bibitem{Wang.2011}
Ko-Hsin~Cindy Wang and Adi Botea.
\newblock {MAPP}: a scalable multi-agent path planning algorithm with
  tractability and completeness guarantees.
\newblock {\em Journal of Artificial Intelligence Research}, 42:55--90, 2011.

\bibitem{yuan2017bot}
Zhe Yuan and Yeming~Yale Gong.
\newblock Bot-in-time delivery for robotic mobile fulfillment systems.
\newblock {\em IEEE Transactions on Engineering Management}, 64(1):83--93,
  2017.

\bibitem{zou2017assignment}
Bipan Zou, Yeming Gong, Xianhao Xu, and Zhe Yuan.
\newblock Assignment rules in robotic mobile fulfilment systems for online
  retailers.
\newblock {\em International Journal of Production Research},
  55(20):6175--6192, 2017.

\end{thebibliography}

\end{document}